\documentclass[12pt]{article}

\usepackage{placeins}
\usepackage{enumitem}
\usepackage{mathtools}
\usepackage{xr}

\usepackage{url}
\usepackage{amsfonts}
\usepackage{amsmath,amssymb,color}
\usepackage{graphicx}
\usepackage{enumitem}
\usepackage{multirow}
\usepackage{tabularx}

\usepackage{bm}
\usepackage{verbatim}
\usepackage{float}
\usepackage[toc,page]{appendix}
\usepackage{apacite}
\usepackage{natbib}
\usepackage{adjustbox}
\usepackage{rotating}
\usepackage{subcaption}
\usepackage{amsthm}
\usepackage{booktabs,dcolumn,caption}
\usepackage{comment}
\usepackage{mathabx}

\usepackage{array}

\usepackage[ruled]{algorithm2e}
\usepackage{amsthm}
\usepackage{hyperref}
\usepackage{cleveref}

\newcolumntype{L}[1]{>{\raggedright\let\newline\\\arraybackslash\hspace{0pt}}p{#1}}
\newcolumntype{C}[1]{>{\centering\let\newline\\\arraybackslash\hspace{0pt}}p{#1}}
\newcolumntype{R}[1]{>{\raggedleft\let\newline\\\arraybackslash\hspace{0pt}}p{#1}}

\usepackage{setspace}

\theoremstyle{plain}

\title{\LARGE  An efficient adaptive dimension selection algorithm for multidimensional probit graded response models}

\author{
  Yu Zhou\textsuperscript{1} \thanks{Email: 52274404005@stu.ecnu.edu.cn}, 
  Yincai Tang\textsuperscript{1}, 
  Bin Lv\textsuperscript{1},
  Meng Gao\textsuperscript{2} \\
  {\small \textsuperscript{1}KLATASDS-MOE} \\
  {\small \textsuperscript{1}School of Statistics, East China Normal University, Shanghai, 200062, China}\\
  {\small \textsuperscript{2}Zhiyuan School of Liberal Arts,Beijing Institute of Petrochemical Technology,Beijing 102617, China}
}

\date{}

\begin{document}
	\setstretch{1.4}
	\maketitle
	
	\begin{abstract}
Multidimensional graded response models (MGRMs) are widely used for analyzing ordinal questionnaire data in psychological and educational assessments. A central challenge in applying these models is determining the number of latent dimensions. Conventional approaches usually fit multiple fixed-dimensional models and select among them using post-hoc criteria such as AIC, BIC, or cross-validation, which can be computationally demanding and ignore uncertainty in dimensionality during estimation.
We develop an adaptive Bayesian dimension selection framework for probit MGRMs. Building on the cumulative shrinkage process, we assign a cumulative ordered spike-and-slab (COSS) prior to the column-specific variances of the item loading matrix. This prior induces increasing shrinkage across latent dimensions, allowing redundant dimensions to be shrunk toward zero while preserving flexibility for active dimensions. Albert--Chib latent response augmentation is used to handle the ordinal probit likelihood, yielding conditionally Gaussian updates for item loadings and latent traits. These updates are combined with Gibbs updates for threshold and shrinkage parameters in an efficient adaptive sampler.
    Simulation studies evaluate the proposed 
  method in terms of dimension recovery, parameter
   estimation accuracy, and computational efficiency, with comparisons to conventional fixed-dimensional estimation and model selection procedures. The results show that the proposed approach accurately recovers the latent structure while avoiding repeated model fitting over multiple candidate dimensions. We further illustrate the method using real psychological assessment data, demonstrating its practical utility for uncovering interpretable latent structures in ordinal item responses.
	\end{abstract}

	\section{Introduction}

Multidimensional item response theory (MIRT) models are now a cornerstone of modern psychometrics, educational testing, and behavioral science. By linking multiple latent traits to observed item responses through a probabilistic measurement framework, MIRT provides a principled way to analyze complex assessment data and to uncover the latent structure underlying human responses \citep{reckase1997past, mcdonald1999test, reckase2009multidimensional}. In many applications, however, responses are often ordinal rather than binary, arising from Likert-type questionnaires, rating scales, or polytomous test items. To model such ordered categorical outcomes, a rich family of polytomous item response models has been developed \citep{masters1982rasch, muraki1992generalized}. Among these models, the graded response model (GRM), originally introduced by \citet{samejima1969estimation} and later systematized by \citet{samejima1997graded}, has become one of the most widely used frameworks. Its appeal lies in the use of ordered threshold parameters, which allow flexible modeling of transitions across response categories. The multidimensional graded response model (MGRM) extends this idea to multiple latent traits and is particularly attractive when several correlated psychological or educational constructs are measured simultaneously \citep{forero2009factor}.

Despite its flexibility and interpretability, the practical use of MGRM raises a fundamental modeling question: how many latent dimensions should be included? This issue is central to the success of multidimensional item response modeling, yet it remains one of the most difficult aspects of implementation. Classical estimation approaches for MIRT models—including marginal maximum likelihood via the EM algorithm \citep{bock1981marginal,bock1988full}, Bayesian MCMC methods \citep{albert1992bayesian, patz1999straightforward, beguin2001mcmc}, and the MH-RM algorithm \citep{cai2010high, chalmers2012mirt, cai2010metropolis}—all require the latent dimension to be fixed in advance. If the dimension is set too small, the model may fail to capture important latent structure and may yield biased parameter estimates. If it is set too large, redundant factors can enter the model, producing overfitting, unstable estimation, and serious interpretability difficulties due to weakly identified or near-zero loadings. In applied research, dimension determination is therefore often carried out using exploratory factor analysis, parallel analysis \citep{horn1965rationale}, or a model comparison strategy based on criteria such as AIC, BIC, DIC, or WAIC \citep{kang2007irt}. While widely used, these strategies can be computationally expensive when multiple high-dimensional MGRMs must be fitted repeatedly, and they may be unreliable in finite samples or sparse ordinal data settings where asymptotic approximations break down \citep{li2025sparse}. Regularized frequentist alternatives based on lasso-type or ridge-type penalties \citep{chen2020regularized} offer another route to sparsity, but they still require tuning-parameter selection, often through cross-validation, which can be burdensome in large-scale psychometric applications.

Bayesian methods provide a coherent alternative for learning latent structure directly from the data \citep{fox2010bayesian, levy2016bayesian}. In particular, shrinkage priors and infinite-dimensional latent factor models have made it possible to infer effective model complexity without relying on repeated model fitting. A key theoretical insight comes from the asymptotic behavior of overfitted mixture models \citep{rousseau2011asymptotic}, which suggests that properly designed sparse priors can empty redundant components automatically. In Bayesian factor analysis, one influential development is the multiplicative gamma process (MGP) prior \citep{bhattacharya2011sparse}, which imposes increasingly strong shrinkage on higher-order factors and thereby encourages parsimonious low-rank representations. This idea has inspired a range of extensions, including sparse Bayesian factor rotation methods \citep{rockova2016fast} and infinite factor models with adaptive shrinkage \citep{fruhwirth2018sparse}.

Building on this line of work, \citet{legramanti2020bayesian} proposed the cumulative shrinkage process, also known as the cumulative ordered spike-and-slab (COSS) prior. The key idea is to impose an ordered sequence of spike probabilities through a stick-breaking construction, so that later latent dimensions are increasingly likely to be shrunk toward zero. Compared with continuous shrinkage priors such as the MGP, the COSS prior offers a more explicit separation between active and inactive factors, making dimension selection more interpretable and often more stable in finite samples. This idea has been extended to generalized infinite factorization models \citep{schiavon2022generalized} and network eigenmodels \citep{loyal2025spike}, but its application to ordinal psychometric models remains limited.

Extending adaptive shrinkage ideas to the MGRM is not straightforward. The ordinal probit likelihood is non-linear and non-conjugate, and once it is combined with a hierarchical shrinkage prior, the posterior distribution becomes analytically intractable. As a result, standard MCMC samplers may mix slowly, especially when the model includes a large number of potential latent dimensions and many loadings are close to zero. Moreover, latent factor models suffer from rotational and sign indeterminacy, which complicates both estimation and interpretation. These issues make adaptive Bayesian dimension selection particularly challenging in the psychometric setting.

To address these difficulties, we propose a fully Bayesian adaptive dimension selection algorithm for multidimensional probit graded response models using the COSS prior. The proposed method places the shrinkage prior on the columns of the loading matrix, so that higher-indexed latent dimensions are progressively discouraged from contributing to the response process. This ordered shrinkage structure provides a soft mechanism for selecting the effective latent dimension while promoting a parsimonious and interpretable factorization of the response model. In this sense, the prior plays a dual role: it regularizes the loading structure and helps align the latent space with decreasing factor importance.

To make posterior inference computationally tractable, we adopt the data augmentation strategy of \citet{albert1993bayesian}. By introducing latent truncated normal variables underlying the ordinal responses, the probit MGRM can be rewritten as a sequence of Gaussian regression updates. This augmentation restores conjugacy for the latent traits, the loadings, and the threshold parameters, allowing us to develop an efficient Gibbs sampling scheme with closed-form conditional updates. Compared with reversible jump MCMC or other trans-dimensional methods \citep{green1995reversible, carlin1995bayesian, stephens2000bayesian,davies2023transport}, the proposed sampler avoids complicated dimension-matching steps and instead learns the model complexity through dynamic truncation during sampling.

The remainder of this paper is organized as follows. Section 2 introduces the multidimensional probit graded response model and the COSS prior specification. Section 3 presents the data augmentation scheme and derives the adaptive Gibbs sampler. Section 4 reports simulation studies that evaluate parameter recovery and dimension-selection accuracy across a range of sample sizes, test lengths, and true dimensionalities. Section 5 illustrates the practical usefulness of the proposed method using real psychometric data. Section 6 concludes the paperand discusses future directions.

\section{Model specification}

Suppose a test consisting of $q$ items is administered to $n$ individuals. Let $Y_{ij}$ denote the random variable representing the ordered categorical response of individual $i$ ($i=1,2,\ldots,n$) to item $j$ ($j=1,2,\ldots,q$), taking values in $\{1, 2, \dots, t_j\}$, where $t_j > 2$ is the number of response categories for item $j$. 
	
We adopt an overfitted working model with a conservatively large number of potential latent dimensions, denoted by $K$. The response of each individual is assumed to be driven by a $K$-dimensional latent trait vector $\boldsymbol{z}_i = (z_{i1}, z_{i2}, \dots, z_{iK})^\top \in \mathbb{R}^{K}$, where redundant dimensions are shrunk toward zero through the cumulative shrinkage prior introduced below.
	
\subsection{ The Multidimensional Probit Graded Response Model}
	
Under the probit specification of the graded response model (GRM), the cumulative probability that individual $i$'s response to item $j$ falls into category $t$ or lower, conditional on the latent trait $\boldsymbol{z}_i$, is modeled using the standard normal cumulative distribution function $\Phi(\cdot)$:
	\begin{equation}\label{eq:cumulative_probit}
		P(Y_{ij} \le t \mid \boldsymbol{z}_i, \boldsymbol{\beta}_j, \boldsymbol{\gamma}_j) = \Phi( \gamma_{j,t}+ \boldsymbol{z}_i^\top \boldsymbol{\beta}_j )  , \quad t = 1,2, \dots, t_j-1,
	\end{equation}
where $\boldsymbol{\beta}_j = (\beta_{j1}, \beta_{j2}, \dots, \beta_{jK})^\top$ is the vector of discrimination parameters (factor loadings) for item $j$, and $\boldsymbol{\gamma}_j = (\gamma_{j,1},\gamma_{j,2}, \dots, \gamma_{j,t_j-1})^\top$ contains the item-specific threshold parameters. 
	
The probability of responding exactly in category $t$ is then given by the difference between adjacent cumulative probabilities:
	\begin{equation}\label{eq:category_prob}
		P(Y_{ij} = t \mid \boldsymbol{z}_i, \boldsymbol{\beta}_j, \boldsymbol{\gamma}_j) = \Phi(\gamma_{j,t} + \boldsymbol{z}_i^\top \boldsymbol{\beta}_j) - \Phi(\gamma_{j,t-1} +\boldsymbol{z}_i^\top \boldsymbol{\beta}_j),
	\end{equation}
where we define the boundary thresholds as $\gamma_{j,0} = -\infty$ and $\gamma_{j,t_j} = +\infty$ to ensure a valid probability mass function. To satisfy the monotonicity of cumulative probabilities, the thresholds must be ordered: $\gamma_{j,0} < \gamma_{j,1} < \gamma_{j,2} < \dots < \gamma_{j,t_j-1} < \gamma_{j,t_j}$.
	
The joint likelihood of the observed response matrix $\mathbf{Y}$ can be expressed as:
	\begin{equation}\label{eq:joint_likelihood}
		p(\mathbf{Y}\mid \mathbf{Z}, \mathbf{B}, \boldsymbol\gamma)
		= \prod_{i=1}^n \prod_{j=1}^q \sum_{t=1}^{t_j}\mathbb{I}(Y_{ij}=t)\left(  \Phi( \gamma_{j,t} + \mathbf{z}_i^\top \boldsymbol{\beta}_j) - \Phi(\gamma_{j,t-1} +\mathbf{z}_i^\top \boldsymbol{\beta}_j) \right),
	\end{equation}
where $\mathbb{I}(\cdot)$ is the indicator function, $\mathbf{Z}$ is the $n \times K$ matrix of latent traits, and $\mathbf{B}$ is the $q \times K$ matrix of factor loadings, $\boldsymbol\gamma$ is the $q \times (t_j-1)$ matrix of threshold parameters.
	
As with all latent variable models, the MGRM faces inherent identifiability challenges. Unrestricted estimation is subject to rotational ambiguity and scale indeterminacy. To address scale indeterminacy, we fix the distribution of the latent traits by assigning a standard multivariate normal prior, $\boldsymbol{z}_i \sim \mathcal{N}(\mathbf{0}, \mathbf{I}_{K})$. 
The proposed cumulative shrinkage prior imposes an ordered structure on the columns of the loading matrix, thereby encouraging a representation in which earlier dimensions explain more variation and later dimensions are increasingly shrunk toward zero. This ordered shrinkage substantially mitigates rotational and column-permutation ambiguity for the purpose of dimension selection.

\subsection{Adaptive Dimension Selection via Cumulative Shrinkage}

A critical challenge in MIRT modeling is the selection of the number of latent dimensions. Let $K_0$ denote the unknown true effective dimension underlying the response data. In practice, $K_0$ is not assumed to be known. A fixed, pre-specified dimension may lead to underfitting if important psychological constructs are omitted, or to overfitting if redundant dimensions are introduced to capture noise. To address this issue, we adopt an overfitted working model with a conservatively large number of potential latent dimensions, denoted by $K$.
	
We then use a cumulative shrinkage prior on the columns of the loading matrix $\mathbf{B}$ to shrink redundant dimensions toward zero and infer the effective number of active dimensions from the data. Specifically, let $\mathbf{B}=(\boldsymbol{\beta}_1,\ldots,\boldsymbol{\beta}_q)^\top$ be the $q\times K$ loading matrix, where the $h$th column corresponds to the loadings on the $h$th potential latent dimension. For each item $j=1,\ldots,q$ and dimension $h=1,\ldots,K$, we assume
	\begin{equation}
		\beta_{jh}\mid \theta_h \sim \mathcal{N}(0,\theta_h),
	\end{equation}
where $\theta_h$ is a column-specific variance parameter controlling the overall magnitude of the $h$th column of $\mathbf{B}$. Small values of $\theta_h$ shrink the corresponding column of loadings toward zero, thereby deactivating the associated latent dimension.
	
The key idea is to impose an ordered shrinkage structure on the sequence $\{\theta_h\}_{h=1}^{K}$, encouraging earlier dimensions to capture the dominant variation in the responses while increasingly shrinking later dimensions toward zero. We formalize this idea using a cumulative ordered spike-and-slab (COSS) prior, building on the cumulative shrinkage process of \citet{legramanti2020bayesian}.
	
\subsubsection{Non-homogeneous Cumulative Shrinkage Process}

For the column-specific variance parameters introduced above, we assign each $\theta_h$ a spike-and-slab prior of the form
\begin{equation}\label{thetah}
	\theta_h \mid \pi_h \sim (1-\pi_h)P_{\mathrm{slab}} + \pi_h P_{\mathrm{spike}},
	\quad h=1,\ldots,K,
\end{equation}
where $P_{\mathrm{slab}}$ is a diffuse distribution that allows active dimensions to have non-negligible loadings, whereas $P_{\mathrm{spike}}$ is concentrated near zero and shrinks redundant dimensions.

To induce increasing shrinkage across dimensions, the spike probabilities are constrained to be nondecreasing:
\[
\pi_1 \leq \pi_2 \leq \cdots \leq \pi_K.
\]
This is achieved through a stick-breaking construction. Specifically, let
\begin{equation}\label{compute_omega}
	\pi_h = \sum_{l=1}^{h} \omega_l,
	\quad
	\omega_l = v_l \prod_{m=1}^{l-1}(1-v_m),
	\quad h=1,\ldots,K,
\end{equation}
where
\[
v_1 \sim \mathrm{Beta}(\kappa,1), 
\quad
v_l \sim \mathrm{Beta}(a,1), \quad l=2,\ldots,K-1,
\quad
v_K=1.
\]
The constraint $v_K=1$ ensures that $\sum_{l=1}^{K}\omega_l=1$ and hence $\pi_K=1$.
The process is \textit{non-homogeneous} in the sense that the stick-breaking variables are not identically distributed. The hyperparameter $\kappa$ controls the prior probability that the first dimension is shrunk, with smaller values of $\kappa$ favoring an active first dimension. The hyperparameter $a$ controls the rate at which shrinkage increases across dimensions; larger values of $a$ place more mass on earlier stick-breaking components, causing the spike probabilities $\pi_h$ to increase more rapidly with $h$.

This construction induces an ordered shrinkage mechanism: early columns of $\mathbf{B}$ are relatively more likely to be associated with the slab component, whereas later columns have increasingly larger prior probabilities of being assigned to the spike component. Consequently, redundant latent dimensions are automatically shrunk toward zero, and the effective dimensionality can be inferred from the posterior distribution of the column-specific variances $\{\theta_h\}_{h=1}^{K}$. When the spike distribution is taken to be a point mass at zero, $P_{\mathrm{spike}}=\delta_0$, this ordered shrinkage mechanism further implies an ordering of the prior probabilities that columns are exactly zero, as formalized below.

\textbf{Proposition 1 (Stochastic Ordering).} \textit{Let $\{ \theta_h \}_{h=1}^K$ follow the prior defined by \eqref{thetah}-\eqref{compute_omega} with $P_{\text{spike}} = \delta_0$. Then for any $\epsilon > 0$ and all $1 \leq h < K$, we have $\Pr(|\theta_{h+1}| \leq \epsilon) > \Pr(|\theta_h| \leq \epsilon)$.}
	
\textit{Proof.} By the law of total probability and $P_{\text{spike}}(|\theta_h| > \epsilon)=0$, we have:
	\[
	\Pr(|\theta_h| > \epsilon) = \mathbb{E}\big[ (1-\pi_h) P_{\mathrm{slab}}(|\theta_h| > \epsilon) \big] = P_{\mathrm{slab}}(|\theta| > \epsilon) \cdot \mathbb{E}(1-\pi_h).
	\]
Based on the stick-breaking construction, $1-\pi_h = \prod_{l=1}^h (1-v_l)$. Since $\{v_l\}$ are independent, and $\mathbb{E}(1-v_1) = \frac{1}{1+\kappa}$, $\mathbb{E}(1-v_l) = \frac{1}{1+a}$ for $l > 1$, we obtain:
	\[
	\mathbb{E}(1-\pi_h) =\mathbb{E}\Big[\prod_{l=1}^h (1-v_l)\Big]= \frac{1}{\kappa+1}\left(\frac{1}{a+1}\right)^{h-1}.
	\]
For $a>0$, $0 < \frac{1}{a+1} < 1$, making $\mathbb{E}(1-\pi_h)$ strictly decreasing with respect to $h$. Consequently, $\Pr(|\theta_{h+1}| > \epsilon) < \Pr(|\theta_h| > \epsilon)$, which directly leads to $\Pr(|\theta_{h+1}| \le \epsilon) > \Pr(|\theta_h| \le \epsilon)$ for all $1 \le h < K$. \hfill $\square$
	
\subsubsection{ Prior Specification for Factor Loadings}

We now apply the proposed COSS framework to the factor loading matrix $\mathbf{B}$. 
For each column $h$, all loading parameters $\beta_{jh}$ share a common column-specific variance parameter $\theta_h$:
\begin{equation}\label{COSS_B}
	\begin{aligned}
		\beta_{jh} \mid \theta_h 
		&\sim \mathcal{N}(0, \theta_h), \quad j = 1, \ldots, q, \\
		\theta_h \mid \pi_h 
		&\sim (1 - \pi_h) \, \mathrm{IG}(a_\theta, b_\theta) 
		+ \pi_h \, \delta_{\theta_0}.
	\end{aligned}
\end{equation}
Here, the slab component is specified as an Inverse-Gamma distribution $\mathrm{IG}(a_\theta,b_\theta)$, while the spike component $\delta_{\theta_0}$ is taken to be a point mass at a very small positive value $\theta_0>0$. 
This choice provides a continuous relaxation of an exact point mass at zero variance, which avoids numerical degeneracy in MCMC sampling while still effectively shrinking loadings in redundant columns toward zero.

After integrating out $\theta_h$, the marginal prior distribution of each loading is
\[
\beta_{jh} \mid \pi_h
\sim 
(1-\pi_h)\, t_{2a_\theta}(0,b_\theta/a_\theta)
+
\pi_h\, \mathcal{N}(0,\theta_0),
\]
where $t_{2a_\theta}(0,b_\theta/a_\theta)$
denotes a Student-\(t\) distribution with degrees of freedom \(2a_\theta\), location 0, and scale squared \(b_\theta / a_\theta\). 

The heavy-tailed slab component protects relevant loadings from excessive shrinkage, whereas the narrow Gaussian spike component concentrates irrelevant loadings near zero. Thus, compared with the exact spike case in Proposition~1, the present continuous 
relaxation yields an analogous ordering in terms of the prior probabilities of near-zero loadings.

\textbf{Corollary 1 (Stochastic Ordering of Loadings).} \textit{Let $\beta_{jh}$ be drawn from the COSS prior \eqref{COSS_B}. For any $\epsilon > 0$, the inequality $\Pr(|\beta_{j,h+1}| \leq \epsilon) > \Pr(|\beta_{jh}| \leq \epsilon)$ holds for all $1 \leq h < K$, provided $\theta_0 < b_\theta / a_\theta$.}
	
	\textit{Proof.}
Let
	\[
	G_0(\epsilon)=\Pr_{\mathcal N(0,\theta_0)}(|X|\le \epsilon)
	\]
and
	\[
	G_1(\epsilon)=\Pr_{t_{2a_\theta}(0,b_\theta/a_\theta)}(|X|\le \epsilon).
	\]
Conditional on \(\pi_h\), the marginal distribution of \(\beta_{jh}\) is
	\[
	\beta_{jh}\mid \pi_h
	\sim
	(1-\pi_h)t_{2a_\theta}(0,b_\theta/a_\theta)
	+
	\pi_h\mathcal N(0,\theta_0).
	\]
Hence
	\[
	\Pr(|\beta_{jh}|\le \epsilon\mid \pi_h)
	=
	G_1(\epsilon)
	+
	\pi_h\{G_0(\epsilon)-G_1(\epsilon)\}.
	\]
Taking expectation with respect to the stick-breaking process gives
	\[
	\Pr(|\beta_{jh}|\le \epsilon)
	=
	G_1(\epsilon)
	+
	\mathbb E(\pi_h)\{G_0(\epsilon)-G_1(\epsilon)\}.
	\]
Therefore,
	\[
	\begin{aligned}
		&\Pr(|\beta_{j,h+1}|\le \epsilon)
		-
		\Pr(|\beta_{jh}|\le \epsilon)
		\\
		&=
		\{\mathbb E(\pi_{h+1})-\mathbb E(\pi_h)\}
		\{G_0(\epsilon)-G_1(\epsilon)\}.
	\end{aligned}
	\]
	
By the stick-breaking construction,
	\[
	\mathbb E(\pi_h)
	=
	1-
	\frac{1}{\kappa+1}
	\left(\frac{1}{a+1}\right)^{h-1},
	\]
and thus
	\[
	\mathbb E(\pi_{h+1})-\mathbb E(\pi_h)
	=
	\frac{1}{\kappa+1}
	\left(\frac{1}{a+1}\right)^{h-1}
	\frac{a}{a+1}
	>0.
	\]
	
It remains to show that \(G_0(\epsilon)>G_1(\epsilon)\). Since
	\[
	G_0(\epsilon)
	=
	2\Phi\left(\frac{\epsilon}{\sqrt{\theta_0}}\right)-1,
	\]
and
	\[
	G_1(\epsilon)
	=
	2T_{2a_\theta}
	\left(
	\frac{\epsilon}{\sqrt{b_\theta/a_\theta}}
	\right)-1,
	\]
where \(T_{2a_\theta}\) denotes the cdf of a standard Student's \(t\) distribution with \(2a_\theta\) degrees of freedom, the condition
	\[
	\theta_0<\frac{b_\theta}{a_\theta}
	\]
	implies
	\[
	\frac{\epsilon}{\sqrt{\theta_0}}
	>
	\frac{\epsilon}{\sqrt{b_\theta/a_\theta}}.
	\]
	Moreover, for any \(x>0\), \(T_\nu(x)<\Phi(x)\). Therefore,
	\[
	\Phi\left(\frac{\epsilon}{\sqrt{\theta_0}}\right)
	>
	\Phi\left(\frac{\epsilon}{\sqrt{b_\theta/a_\theta}}\right)
	>
	T_{2a_\theta}
	\left(
	\frac{\epsilon}{\sqrt{b_\theta/a_\theta}}
	\right),
	\]
	which implies \(G_0(\epsilon)>G_1(\epsilon)\). Consequently,
	\[
	\Pr(|\beta_{j,h+1}|\le \epsilon)
	>
	\Pr(|\beta_{jh}|\le \epsilon).
	\]
	This completes the proof. \(\square\)

Through the stochastic ordering induced by the COSS prior, later columns of $\mathbf{B}$ are progressively assigned stronger shrinkage toward zero. In the probit GRM setting, this ordered shrinkage encourages the retained latent traits to be arranged according to decreasing explanatory relevance, while redundant latent dimensions are shrunk toward zero. It therefore provides a shrinkage-based mechanism for dimension selection and enables posterior inference on the effective number of latent traits.

\section{Posterior Inference via Adaptive Gibbs Sampling}

In this section, we develop an efficient Markov chain Monte Carlo (MCMC) algorithm for posterior inference in the multidimensional probit graded response model equipped with the cumulative ordered spike-and-slab (COSS) prior. The main challenges in posterior 
sampling arise from two sources: first, the nonconjugate form of the ordinal probit likelihood before data augmentation; and second, the complex hierarchical structure 
of the COSS prior.

To address these challenges, we construct a Gibbs sampling algorithm based on data augmentation and an adaptive truncation scheme. The data augmentation step restores conditional conjugacy for the ordinal probit component, while the adaptive truncation 
scheme allows the sampler to focus computation on the relevant latent dimensions. Together, these components enable efficient posterior exploration of the model parameters 
and the effective number of latent traits. We first present the sampler under a fixed truncation level $K$, and then introduce an adaptive scheme that dynamically adjusts $K$ during the MCMC iterations to improve computational efficiency.

\subsection{Data Augmentation and Parameter Updates} 
	
To overcome the non-conjugacy of the probit MGRM, we adopt the latent variable data augmentation strategy proposed by \citet{albert1993bayesian}. Specifically, we introduce a continuous latent response variable $Y_{ij}^*$ for each observed ordinal response $Y_{ij}$. Conditional on the latent traits $\boldsymbol{z}_i$, the factor loadings $\boldsymbol{\beta}_j$, and the observed category $Y_{ij} = t$, $Y_{ij}^*$ is drawn from a truncated normal distribution:
	\begin{equation} \label{eq:Ystar}
		Y_{ij}^* \mid \boldsymbol{z}_i, \boldsymbol{\beta}_j, Y_{ij} = t \sim \mathcal{N}(-\boldsymbol{z}_i^\top \boldsymbol{\beta}_j, 1)\mathbb{I}(\gamma_{j, t-1} < Y_{ij}^* \leq \gamma_{j, t}),
	\end{equation}
where $\mathbb{I}(\cdot)$ is the indicator function. The introduction of $Y_{ij}^*$ reformulates the probit MGRM into a standard Bayesian linear regression model, $Y_{ij}^* = -\boldsymbol{z}_i^\top \boldsymbol{\beta}_j + \epsilon_{ij}$, with $\epsilon_{ij} \sim \mathcal{N}(0,1)$, thereby rendering the conditional posterior distributions of $\mathbf{Z}$ and $\mathbf{B}$ analytically tractable.
	
Given the augmented data $\mathbf{Y}^* = \{Y_{ij}^*\}$, the full conditional posterior distribution for the factor loading vector $\boldsymbol{\beta}_j$ of item $j$ follows a multivariate normal distribution:
	\begin{equation}\label{eq:B_update}
		\boldsymbol{\beta}_j \mid \text{rest} \sim \mathcal{N}(\boldsymbol{\mu}_{\beta_j}, \boldsymbol{\Sigma}_{\beta_j}),
	\end{equation}
where the posterior covariance and mean are given by
	\[
	\boldsymbol{\Sigma}_{\beta_j} = \left(\boldsymbol{\Sigma}_B^{-1} + \mathbf{Z}^\top \mathbf{Z}\right)^{-1} \quad \text{and} \quad \boldsymbol{\mu}_{\beta_j} = \boldsymbol{\Sigma}_{\beta_j} (-\mathbf{Z}^\top \mathbf{Y}^{*}_{\cdot j}),
	\]
respectively. Here, $\mathbf{Y}^{*}_{\cdot j} = (Y_{1j}^*, Y_{2j}^*, \dots, Y_{nj}^*)^\top$, and $\boldsymbol{\Sigma}_B = \text{diag}(\theta_1, \dots, \theta_K)$ represents the prior covariance matrix governed by the COSS prior. To ensure strict identifiability, lower-triangular constraints are imposed on $\mathbf{B}$ during the sampling process.
	
Similarly, the full conditional posterior distribution for the latent trait vector $\boldsymbol{z}_i$ of individual $i$ is:
	\begin{equation}\label{eq:Z_update}
		\boldsymbol{z}_i \mid \text{rest} \sim \mathcal{N}(\boldsymbol{\mu}_{z_i}, \boldsymbol{\Sigma}_{z_i}),
	\end{equation}
with the covariance and mean given by
	\[
	\boldsymbol{\Sigma}_{z_i} = \left( \mathbf{I}_K + \mathbf{B}^\top \mathbf{B} \right)^{-1} \quad \text{and} \quad \boldsymbol{\mu}_{z_i} = \boldsymbol{\Sigma}_{z_i} (-\mathbf{B}^\top \mathbf{Y}^{*}_{i\cdot}),
	\]
where $\mathbf{Y}^{*}_{i\cdot} = (Y_{i1}^*, Y_{i2}^*, \dots, Y_{iq}^*)^\top$. Since a standard normal prior $\mathcal{N}(\mathbf{0}, \mathbf{I}_K)$ is assigned to $\boldsymbol{z}_i$ to anchor the scale of the latent space, the prior precision matrix reduces to the identity matrix $\mathbf{I}_K$.
	
For the threshold parameters $\gamma_{j,t}$, assuming a diffuse prior bounded by appropriate constraints, the full conditional distribution is uniform over a restricted interval:
	\begin{equation}\label{eq:gamma_update}
		\gamma_{j,t} \mid \text{rest} \sim \text{Uniform} (\Omega_L, \Omega_U), 
	\end{equation}
where the lower and upper bounds are determined by the maximum and minimum values of the augmented data within adjacent categories, combined with the order constraints from neighboring thresholds:
	$
	\Omega_L = \max \left( \max_{i:Y_{ij}=t} Y_{ij}^{*}, \,\gamma_{j,t-1} \right), 
	\Omega_U = \min \left( \min_{i:Y_{ij}=t+1} Y_{ij}^{*}, \, \gamma_{j,t+1} \right).
$
	
	\subsection{Updating the COSS Prior Parameters} 
To facilitate posterior computation under the COSS prior, we introduce a latent allocation indicator 
$\rho_h \in \{1,2,\ldots,K\}$ for each dimension $h$, with
$
\Pr(\rho_h=l\mid \boldsymbol{\omega})=\omega_l, l=1,\ldots,K.
$
Recall that the spike probability is 
$
\pi_h=\sum_{l=1}^h \omega_l.
$
It follows that
$
\Pr(\rho_h\le h\mid \boldsymbol{\omega})
=
\sum_{l=1}^h \omega_l
=
\pi_h.
$
Therefore, the event $\rho_h\le h$ corresponds to assigning the $h$th dimension to the spike component, whereas $\rho_h>h$ corresponds to assigning it to the slab component. This yields a latent-variable representation that is marginally equivalent to the original spike-and-slab prior.
Specifically,
\begin{equation}
	\theta_h\mid \rho_h \sim
	\begin{cases}
		\delta_{\theta_0}, & \rho_h\le h,\\
		\operatorname{IG}(a_\theta,b_\theta), & \rho_h>h,
	\end{cases}
	\label{eq:theta_csp_prior}
\end{equation}
Equivalently,
\begin{equation}
	\theta_h \mid \rho_h \sim 
	\mathbb{I}(\rho_h \leq h) \, \delta_{\theta_0} 
	+ 
	\left\{1 - \mathbb{I}(\rho_h \leq h)\right\} \, \operatorname{IG}(a_\theta, b_\theta).
\end{equation}
Marginalizing out independent indicators $\rho_h$ recovers
$
\theta_h\mid \pi_h
\sim
\pi_h\delta_{\theta_0}
+
(1-\pi_h)\operatorname{IG}(a_\theta,b_\theta),
$
which is the original COSS spike-and-slab prior. The number of active dimensions can then be counted as
$
K^*=\sum_{h=1}^K \mathbb{I}(\rho_h>h).
$

The latent indicators $\rho_h$ are updated by sampling from their conditional posterior distribution. 
For each $h$, by Bayes' rule,
$
\Pr(\rho_h=l \mid \boldsymbol{\beta}_{\cdot h},\boldsymbol{\omega})
\propto
\Pr(\rho_h=l\mid \boldsymbol{\omega})
p(\boldsymbol{\beta}_{\cdot h}\mid \rho_h=l),
$
where $\Pr(\rho_h=l\mid \boldsymbol{\omega})=\omega_l$. 
It remains to derive the marginal density 
$p(\boldsymbol{\beta}_{\cdot h}\mid \rho_h=l)$.

Recall that
$
\boldsymbol{\beta}_{\cdot h}\mid \theta_h
\sim
\mathcal{N}_q(\mathbf{0},\theta_h\mathbf{I}_q)
$
and ~\eqref{eq:theta_csp_prior},
when $l\le h$, the event $\rho_h=l$ implies that $\theta_h=\theta_0$. Hence,
$
p(\boldsymbol{\beta}_{\cdot h}\mid \rho_h=l)
=
\mathcal{N}_q(\boldsymbol{\beta}_{\cdot h};\mathbf{0},\theta_0\mathbf{I}_q), l\le h.
$
When $l>h$, $\theta_h$ follows the inverse-gamma slab prior. Therefore,
\[
\begin{aligned}
	p(\boldsymbol{\beta}_{\cdot h}\mid \rho_h=l)
	&=
	\int
	p(\boldsymbol{\beta}_{\cdot h}\mid \theta_h)
	p(\theta_h\mid \rho_h=l)
	\,d\theta_h  \\
	&=
	\int
	\mathcal{N}_q(\boldsymbol{\beta}_{\cdot h};\mathbf{0},\theta_h\mathbf{I}_q)
	\operatorname{IG}(\theta_h;a_\theta,b_\theta)
	\,d\theta_h.
\end{aligned}
\]
Using the Gaussian--inverse-gamma conjugacy, this integral yields a $q$-variate Student-$t$ density,
$
p(\boldsymbol{\beta}_{\cdot h}\mid \rho_h=l)
=
t_{2a_\theta}
\left(
\boldsymbol{\beta}_{\cdot h};
\mathbf{0},
\frac{b_\theta}{a_\theta}\mathbf{I}_q
\right), l>h.
$
Combining these results, the conditional posterior probabilities of $\rho_h$ are given by
\begin{equation}\label{eq:rho_update}
	\Pr(\rho_h = l \mid \text{rest}) \propto 
	\begin{cases} 
		\omega_l \,
		\mathcal{N}_q(\boldsymbol{\beta}_{\cdot h}; \mathbf{0}, \theta_0 \mathbf{I}_q),
		& \text{if } l \leq h, \\[6pt]
		\omega_l \,
		t_{2a_\theta}(\boldsymbol{\beta}_{\cdot h}; \mathbf{0}, (b_\theta / a_\theta) \mathbf{I}_q),
		& \text{if } l > h.
	\end{cases}   
\end{equation}
Here,
$\mathcal{N}_q(\cdot;\mathbf{0},\theta_0\mathbf{I}_q)$ denotes the $q$-variate Gaussian density, and
$t_{2a_\theta}(\cdot;\mathbf{0},(b_\theta/a_\theta)\mathbf{I}_q)$ denotes the $q$-variate Student-$t$ density with $2a_\theta$ degrees of freedom, location $\mathbf{0}$, and scale matrix $(b_\theta/a_\theta)\mathbf{I}_q$.

Furthermore, the stick-breaking weights are given by
$
\omega_l = v_l\prod_{m<l}(1-v_m), l=1,\ldots,K,
$
with $v_K=1$ under the truncated representation.
Given the stick-breaking weights $\boldsymbol{\omega}$, the latent indicators
$\rho_1,\ldots,\rho_K$ are conditionally independent, with
$
\Pr(\rho_j=l\mid \boldsymbol{\omega})=\omega_l.
$
Since $\boldsymbol{\omega}$ is determined by the stick-breaking variables
$\boldsymbol v$, this can equivalently be written as
$
\Pr(\rho_j=l\mid \boldsymbol v)=\omega_l.
$
Therefore, the joint conditional probability of the indicators is
$
p(\rho_1,\ldots,\rho_K\mid \boldsymbol v)
=
\prod_{j=1}^K p(\rho_j\mid \boldsymbol v)
=
\prod_{j=1}^K \omega_{\rho_j}.
$

For a fixed stick-breaking variable $v_l$, each indicator $\rho_j$ contributes
a factor $v_l$ if $\rho_j=l$, a factor $(1-v_l)$ if $\rho_j>l$, and no factor
involving $v_l$ if $\rho_j<l$. Hence,
$
p(\rho_1,\ldots,\rho_K\mid \boldsymbol v)
\propto
v_l^{\sum_{j=1}^K \mathbb{I}(\rho_j=l)}
(1-v_l)^{\sum_{j=1}^K \mathbb{I}(\rho_j>l)}.
$
Combining this likelihood with the Beta priors gives the Gibbs updates. For
$v_1\sim\operatorname{Beta}(\kappa,1)$,
\begin{equation} \label{eq:v1_update}
	v_1 \mid \text{rest}
	\sim
	\operatorname{Beta}
	\left(
	\kappa + \sum_{j=1}^K \mathbb{I}(\rho_j = 1),
	\;
	1 + \sum_{j=1}^K \mathbb{I}(\rho_j > 1)
	\right).
\end{equation}
For $l=2,\ldots,K-1$, if $v_l\sim\operatorname{Beta}(a,1)$, then
\begin{equation} \label{eq:vh_update}
	v_l \mid \text{rest}
	\sim
	\operatorname{Beta}
	\left(
	a + \sum_{j=1}^K \mathbb{I}(\rho_j = l),
	\;
	1 + \sum_{j=1}^K \mathbb{I}(\rho_j > l)
	\right).
\end{equation}

Finally, given the updated allocation variables $\rho_h$, the variance
parameters $\theta_h$ are updated according to the spike--slab structure. If $\rho_h \le h$, $\theta_h$ is deterministically set to the near-zero constant $\theta_0$. Conversely, if $\rho_h > h$, the inverse-gamma prior remains conjugate to the Gaussian likelihood of $\boldsymbol{\beta}_{\cdot h}$, yielding:
\begin{equation}\label{eq:theta_slab_update}
	\theta_h \mid \rho_h > h, \text{rest} \sim \text{IG}\left( a_\theta + \frac{q}{2}, \; b_\theta + \frac{1}{2} \|\boldsymbol{\beta}_{\cdot h}\|_2^2 \right).
\end{equation}
In summary, given the current state
$(\mathbf{Z}, \boldsymbol{\gamma}, \mathbf{B}, \boldsymbol{\theta},
\boldsymbol{\rho}, \mathbf v)$, with $\boldsymbol{\omega}=(\omega_1,\omega_2,\ldots,\omega_K)$ computed from
$\mathbf v$, the exact Gibbs sampler proceeds by sequentially sampling or
updating each block from the closed-form full conditional distributions
\eqref{eq:Ystar}, \eqref{eq:B_update}, \eqref{eq:Z_update},
\eqref{eq:gamma_update}, \eqref{eq:rho_update}, \eqref{eq:v1_update},
\eqref{eq:vh_update}, and \eqref{eq:theta_slab_update}.

\subsection{Adaptive Truncation Strategy}

The exact Gibbs sampler described above is derived under a fixed truncation level \(K\). Under the finite cumulative shrinkage construction, the last component is reserved as a terminal spike component. Therefore, at most \(K-1\) dimensions can be active. Although one may choose a conservative truncation level, for example \(K=q+1\), such a choice is often computationally inefficient. In applications with a large number of items, the true number of latent traits is typically much smaller than \(q\), and most columns of the loading matrix are expected to be shrunk to the spike component.

To avoid repeatedly updating unnecessary spike columns, we adopt an adaptive truncation strategy following
\citet{bhattacharya2011sparse,legramanti2020bayesian}. The idea is to adjust the truncation level during the MCMC run according to the current number of active dimensions. At iteration \(t\), after a pre-specified burn-in period \(\bar t\),
an adaptation step is attempted with probability
$
p(t)=\exp(\alpha_0+\alpha_1 t),
\alpha_0\le 0, \alpha_1<0.
$
Since this probability decreases to zero as \(t\) increases, the adaptation becomes less frequent along the chain, which is in line with the diminishing adaptation condition of \citet{roberts2007coupling}.

Given the current truncation level \(K\), define the number of active dimensions as
$
K^*
=
\sum_{h=1}^{K}
\mathbb{I}(\rho_h>h).
$
Here dimension \(h\) is active when \(\rho_h>h\), and it is assigned to the spike
component when \(\rho_h\le h\). Since the last component is the
terminal spike component, we have \(K^*\le K-1\).
If \(K^*<K-1\), then the current truncation level contains redundant inactive
dimensions. In this case, the inactive columns of \(\mathbf{B}\), together with
their associated parameters, are removed from the sampler, and the truncation
level is reset to
$
K_{\mathrm{new}}=K^*+1.
$
The extra component is retained as the terminal spike component. On the other
hand, if \(K^*=K-1\), then all available non-terminal dimensions are active,
suggesting that the current truncation level may be insufficient. The sampler
then expands the model by setting
$
K_{\mathrm{new}}=K+1,
$
and initializing the newly introduced final component from its corresponding
prior distribution.

This adaptive scheme allows the sampler to keep only the dimensions that are
currently needed, while still leaving one terminal spike component for possible
future expansion. As a result, it avoids the computational burden caused by an
overly conservative fixed truncation level, without preventing the posterior from
exploring larger latent dimensions when supported by the data. The adaptive procedure proceeds as follows:
\begin{enumerate}
	\item \textbf{Evaluation.}
Compute the current number of active dimensions
	$
	K^*
	=
	\sum_{h=1}^{K}
	\mathbb{I}(\rho_h>h).
	$
	
\item \textbf{Contraction step.}
If $K^*<K-1,$ then the current truncation level is larger than needed. In this case, all
inactive columns of \(\mathbf{B}\), together with their associated parameters, are removed from the sampler. The truncation level is then reset to
	$
	K_{\mathrm{new}}=K^*+1.
	$
The additional final component is kept as the terminal spike component, so that the sampler still has one inactive dimension available for subsequent exploration.
	
\item \textbf{Expansion step.}
If $K^*=K-1,$ then all currently available non-terminal dimensions are active. This suggests that the current truncation level may be too small. In this case, the sampler increases the truncation level by one: $K_{\mathrm{new}}=K+1.$
A new final column is added to \(\mathbf{B}\), and its associated parameters are sampled from the corresponding prior distributions. This newly added component is initialized as the terminal spike component.
\end{enumerate}

Combining the Gibbs updates derived above with the adaptive truncation strategy, we obtain the adaptive Gibbs sampler summarized in Algorithm~\ref{alg:adaptive_gibbs}.

\begin{algorithm}[htbp]
	\caption{Adaptive Gibbs sampler for the probit GRM}
	\label{alg:adaptive_gibbs}
	
	\KwIn{Total iterations \(T\), burn in \(T_0\), initial truncation level \(K\), adaptation parameters \(\alpha_0,\alpha_1\), adaptation starting iteration \(\bar t\)}
	\KwOut{Posterior samples of \(\mathbf{Z}, \mathbf{B}, \boldsymbol{\gamma}\), and the estimated active dimension indicators}
	
	Initialize \(K\), \(\mathbf{Z}\), \(\mathbf{B}\), \(\boldsymbol{\gamma}\), \(\boldsymbol{\theta}\), \(\boldsymbol{\rho} \);
	
	\For{\(t=1,\ldots,T\)}{
		
		Sample augmented responses \(Y_{ij}^*\) from the truncated normal distribution in \eqref{eq:Ystar}\;
		
		Update item loading vectors \(\boldsymbol{\beta}_j\), \(j=1,\ldots,q\), using \eqref{eq:B_update}
		\;
		
		Update latent trait vectors \(\boldsymbol{z}_i\), \(i=1,\ldots,n\), using \eqref{eq:Z_update}\;
		
		Update threshold parameters \(\gamma_{j,t}\) using \eqref{eq:gamma_update}\;
		
		Update allocation indicators \(\rho_h\), \(h=1,\ldots,K\), using \eqref{eq:rho_update}\;
		
		Update stick-breaking variables \(v_h\), \(h=1,\ldots,K-1\), using \eqref{eq:v1_update}--\eqref{eq:vh_update}, and compute \(\boldsymbol{\omega}\)\;
		
		Update variance parameters \(\theta_h\), \(h=1,\ldots,K\), using \eqref{eq:theta_slab_update}; set \(\theta_h=\theta_0\) when \(\rho_h\le h\)\;
		
	\If{\(t>\bar t\)}{
		With probability \(p(t)=\exp(\alpha_0+\alpha_1 t)\), compute
		$
		K^*
		=
		\sum_{h=1}^{K}
		\mathbb{I}(\rho_h>h)
		$\;
		
		\If{\(K^*<K-1\)}{
			Remove the inactive columns of \(\mathbf{B}\), namely those assigned to
			the spike component with \(\rho_h\le h\), together with their associated
			parameters in \(\boldsymbol{\theta}\), \(\boldsymbol{\rho}\),
			\(\mathbf{Z}\), and \(\boldsymbol{\omega}\)\;
			
			Set \(K_{\mathrm{new}}=K^*+1\)\;
			
			Add a final component to \(\mathbf{B}\), \(\boldsymbol{\theta}\), \(\boldsymbol{\rho}\),
			\(\mathbf{Z}\), and \(\boldsymbol{\omega}\) sampled 
			from the corresponding prior distributions\;
			
			Set \(K\leftarrow K_{\mathrm{new}}\)\;
			
		}
		
		\ElseIf{\(K^*\ge K-1\)}{
			Set \(K_{\mathrm{new}}=K+1\)\;
			
			Add a final column sampled from the spike to \(\mathbf{B}\), together with the associated parameters in \(\boldsymbol{\theta}\), \(\boldsymbol{\rho}\),
			\(\mathbf{Z}\), and \(\boldsymbol{\omega}\) sampled
			from the corresponding prior distributions\;
			
			Set \(K\leftarrow K_{\mathrm{new}}\)\;
			
		}
	}
	}
	
\end{algorithm}

\section{Simulation Study}
	
\subsection{Performance of the Adaptive Probit GRM} 
	
In this simulation study, we evaluate the performance of the proposed adaptive 
approach in terms of parameter recovery, latent structure reconstruction, and 
adaptive dimension selection under varying sample sizes ($n$), test lengths 
($q$), and true latent dimensionalities ($K_0 \in \{2,3,4\}$). For each 
simulation condition, we generated 25 independent datasets.

The observed ordinal responses $\mathbf{Y}$ were generated from the probit 
graded response model. To satisfy the strict monotonicity constraint on the 
threshold parameters, the item-specific intercept parameters 
$\boldsymbol{\gamma}$ were independently generated from non-overlapping uniform 
distributions. The true latent traits were generated as 
$\boldsymbol{z}_i \sim \mathcal{N}(\mathbf{0}, \mathbf{I}_{K_0})$. For the true 
loading matrix $\mathbf{B}_0 \in \mathbb{R}^{J \times K_0}$, we used a simple 
structure design: each item loaded on exactly one latent dimension, and the 
nonzero loading was independently drawn from $\mathcal{U}(0.25, 1.25)$. This 
design ensures that each item measures a single latent trait.

In the MCMC implementation, the maximum truncation level was set to 
$K=8$. For the COSS prior imposed on the loading matrix, the slab 
hyperparameters were set to $a_{\theta_B}=b_{\theta_B}=2$, the stick-breaking 
parameter was set to $a=2$, and the spike variance was fixed at 
$\theta_0=0.05$. The Gibbs sampler was run for 20,000 iterations, with the first 
10,000 iterations discarded as burn-in. A thinning interval of 10 was applied to 
the post-burn-in samples. The adaptive truncation mechanism was activated after 
$\bar{t}=500$ iterations, with the decaying probability parameters set to 
$(\eta_0,\eta_1)=(-1,-5\times 10^{-4})$.

To evaluate parameter recovery while accounting for the rotational invariance of
	latent variable models, we used normalized Frobenius errors based on the induced
	inner-product matrices:
	\[
	\Delta_Z =
	\frac{\|\hat{\mathbf{Z}}\hat{\mathbf{Z}}^\top
		- \mathbf{Z}_0\mathbf{Z}_0^\top\|_F}{n},
	\qquad
	\Delta_B =
	\frac{\|\hat{\mathbf{B}}^\top\hat{\mathbf{B}}
		- \mathbf{B}_0^\top\mathbf{B}_0\|_F}{q}.
	\]
These criteria are invariant to orthogonal rotations, column permutations, and
sign changes of the latent factors. The accuracy of threshold estimation was measured by the root mean squared error, denoted by \(\gamma_{\rm rmse}\).
	
We also report global bias measures for the posterior mean estimates. For the threshold parameters, bias was computed as the average entrywise difference between \(\hat{\boldsymbol{\gamma}}\) and \(\boldsymbol{\gamma}_0\). For the loading matrix and latent scores, bias was computed as the average entrywise difference between the estimated and true inner-product matrices,
	\(\hat{\mathbf{B}}^\top\hat{\mathbf{B}} - \mathbf{B}_0^\top\mathbf{B}_0\) and
	\(\hat{\mathbf{Z}}\hat{\mathbf{Z}}^\top - \mathbf{Z}_0\mathbf{Z}_0^\top\),
respectively.
	
For dimension recovery, the estimated latent dimension was defined as the posterior mode of the number of active factors:
	\[
	\hat{K} = \arg\max_l \Pr(K^*=l \mid \mathbf{Y}).
	\]
We report the accuracy rate, denoted by Acc, which is the proportion of replications satisfying \(\hat{K}=K_0\), and the mean absolute bias, denoted by MAB, computed conditional on the replications in which \(\hat{K}\neq K_0\).

Table~\ref{tab:sim_grm} summarizes the simulation results for the proposed probit GRM under different combinations of sample size, test length, and true latent dimensionality. Overall, the proposed Gibbs sampler provides accurate parameter recovery, and its performance improves as the amount of information in the data increases. In particular, the threshold RMSE $\gamma_{\rm rmse}$ 
decreases substantially as the sample size increases from $n=200$ to $n=1000$. Similarly, the loading error $\Delta_B$ and the latent trait error $\Delta_Z$ generally decrease as either $n$ or $q$ increases. These trends provide empirical support for the consistency behavior of the proposed estimation procedure.

The global bias measures are close to zero across most settings, suggesting that the COSS prior induces sparsity without introducing substantial systematic bias for the active parameters. This is particularly important because shrinkage priors may, in principle, over-shrink nonzero loadings. The small values of 
$B_{\rm bias}$ observed in the simulations indicate that the proposed prior achieves a favorable balance between sparsity and estimation accuracy.

For dimension recovery, the proposed method performs very well in the lower-dimensional cases. When the true dimension is $K_0=2$, the correct dimension is selected in all replications across all combinations of $n$ and $q$. For $K_0=3$, the accuracy is also high, reaching $100\%$ in almost all settings once 
the test length is at least $q=60$.

The most challenging scenario occurs when the true latent dimension is $K_0=4$, especially under small sample size and short test length. For example, when $n=200$ and $q=40$, the accuracy drops to $16\%$. This behavior is expected, as each latent dimension is measured by only about $10$ items on average, and the 
available ordinal information is insufficient to fully overcome the shrinkage toward a more parsimonious model. As either the test length or sample size increases, however, the dimension recovery improves markedly. For example, when $n=1000$ and $q=100$, the accuracy reaches $96\%$. Moreover, when dimension 
selection is incorrect, the conditional mean absolute bias is mostly between one and two, indicating that the selected dimension typically remains close to the true dimension rather than exhibiting large selection errors.

\begin{table}[htbp]
	\centering
	\scriptsize
	\setlength{\tabcolsep}{3.5pt}
	\caption{Simulation results for the proposed adaptive probit graded response model under different sample sizes, test lengths, and true latent dimensions.}
	\label{tab:sim_grm}
	\resizebox{\textwidth}{!}{
		\begin{tabular}{lcccccccccccc}
			\toprule
			& \multicolumn{4}{c}{$n=200$} 
			& \multicolumn{4}{c}{$n=500$} 
			& \multicolumn{4}{c}{$n=1000$} \\
			\cmidrule(lr){2-5} \cmidrule(lr){6-9} \cmidrule(lr){10-13}
			Metric 
			& $q=40$ & $q=60$ & $q=80$ & $q=100$
			& $q=40$ & $q=60$ & $q=80$ & $q=100$
			& $q=40$ & $q=60$ & $q=80$ & $q=100$ \\
			\midrule
			
			\multicolumn{13}{l}{\textit{True dimension: $K_0=2$}} \\
			$\gamma_{\rm rmse}$ 
			& 0.168 & 0.169 & 0.163 & 0.166 
			& 0.116 & 0.115 & 0.112 & 0.111 
			& 0.085 & 0.082 & 0.081 & 0.081 \\
			$\Delta_B$ 
			& 0.125 & 0.115 & 0.116 & 0.121 
			& 0.080 & 0.074 & 0.077 & 0.072 
			& 0.056 & 0.054 & 0.055 & 0.053 \\
			$\Delta_Z$ 
			& 0.633 & 0.537 & 0.466 & 0.442 
			& 0.610 & 0.520 & 0.452 & 0.412 
			& 0.595 & 0.510 & 0.436 & 0.401 \\
			Acc 
			& 100\% & 100\% & 100\% & 100\%
			& 100\% & 100\% & 100\% & 100\%
			& 100\% & 100\% & 100\% & 100\% \\
			MAB 
			& 0.000 & 0.000 & 0.000 & 0.000
			& 0.000 & 0.000 & 0.000 & 0.000
			& 0.000 & 0.000 & 0.000 & 0.000 \\
			$\gamma_{\rm bias}$ 
			& -0.007 & -0.005 & -0.007 & 0.003
			& -0.003 & 0.002 & -0.005 & -0.003
			& -0.001 & -0.003 & 0.002 & -0.001 \\
			$B_{\rm bias}$ 
			& 0.015 & 0.005 & 0.006 & 0.014
			& 0.010 & 0.004 & 0.005 & -0.001
			& 0.007 & 0.006 & 0.002 & 0.007 \\
			$Z_{\rm bias}$ 
			& -0.003 & -0.002 & 0.001 & -0.005
			& -0.001 & 0.000 & 0.000 & -0.001
			& 0.000 & 0.000 & 0.000 & 0.000 \\
			
			\addlinespace[0.6em]
			\multicolumn{13}{l}{\textit{True dimension: $K_0=3$}} \\
			$\gamma_{rmse}$ 
			& $0.167$ & $0.167$ & $0.168$ &$0.167$ & $0.115$ & $0.111$ & $0.114$ & $0.112$ & $0.085$ & $0.082$ & $0.084$ & $0.083$ 
			\\
			$\Delta B$ 
			& $0.145$ & $0.123$ & $0.120$ & $0.117$ & $0.083$ & $0.076$ & $0.075$ & $0.071$ & $0.061$ & $0.054$ & $0.053$ & $0.051$ 
			\\
			$\Delta Z$ 
			& $0.955$ & $0.780$ & $0.678$ & $0.634$ & $0.896$ & $0.738$ & $0.674$ & $0.607$ & $0.888$ & $0.738$ & $0.658$ & $0.592$ 
			\\
			$\textbf{Acc}$ 
			& $72\%$ & $100\%$ & $100\%$ & $100\%$ & $96\%$ & $100\%$ & $100\%$ & $100\%$ & $100\%$ & $100\%$ & $100\%$ & $100\%$
			\\
			$\textbf{MAB}$
		 & $1.143$ & $0.000$ & $0.000$ & $0.000$ & $1.000$ & $0.000$ & $0.000$ & $0.000$ & $1.000$ & $0.000$ & $0.000$ & $0.000$ 
		 \\
			$\gamma_{bias}$ 
			& $0.003$ & $-0.014$ & $-0.001$ & $0.003$ & $-0.006$ & $-0.001$ & $-0.003$ & $0.002$ & $-0.002$ & $-0.004$ & $0.002$ & $-0.006$ \\
			$B_{bias}$ 
			& $-0.006$ & $0.012$ & $0.008$ & $0.008$ & $0.002$ & $0.001$ & $0.007$ & $0.004$ & $-0.003$ & $0.004$ & $0.002$ & $0.004$
			\\
			$Z_{bias}$ & $-0.000$ & $-0.002$ & $0.000$ & $-0.004$ & $0.001$ & $0.001$ & $-0.000$ & $0.001$ & $-0.001$ & $0.001$ & $0.001$ & $-0.000$ 
			\\
			
			\addlinespace[0.6em]
			\multicolumn{13}{l}{\textit{True dimension: $K_0=4$}} \\
		$\gamma_{rmse}$ 
		& $0.187$ & $0.175$ & $0.171$ & $0.166$ & $0.132$ & $0.118$ & $0.115$ & $0.113$ & $0.099$ & $0.087$ & $0.085$ & $0.085$ 
		\\
		$\Delta B$ & $0.185$ & $0.166$ & $0.141$ & $0.129$ & $0.142$ & $0.106$ & $0.089$ & $0.078$ & $0.117$ & $0.064$ & $0.056$ & $0.054$ 
		\\
		$\Delta Z$ 
		& $1.487$ & $1.231$ & $1.021$ & $0.910$ & $1.388$ & $1.108$ & $0.944$ & $0.859$ & $1.297$ & $0.985$ & $0.873$ & $0.787$ 
		\\
		$\textbf{Acc}$ 
		& $16\%$ & $32\%$ & $60\%$ & $64\%$ & $28\%$ & $56\%$ & $72\%$ & $84\%$ & $36\%$ & $80\%$ & $92\%$ & $96\%$ 
		\\
		$\textbf{MAB}$
		 & $2.000$ & $1.529$ & $1.500$ & $1.000$ & $1.789$ & $1.182$ & $1.286$ & $1.250$ & $1.562$ & $1.000$ & $1.000$ & $1.000$ 
		\\
		$\gamma_{bias}$
		 & $-0.002$ & $0.001$ & $-0.001$ & $-0.003$ & $0.005$ & $-0.004$ & $0.003$ & $0.002$ & $-0.000$ & $-0.003$ & $0.004$ & $0.001$
		 \\
		$B_{bias}$ 
		& $-0.045$ & $-0.036$ & $-0.014$ & $-0.008$ & $-0.036$ & $-0.012$ & $-0.009$ & $0.001$ & $-0.027$ & $-0.002$ & $0.001$ & $0.002$ 
		\\
		$Z_{bias}$ 
		& $-0.005$ & $-0.004$ & $-0.003$ & $-0.000$ & $-0.002$ & $-0.004$ & $-0.002$ & $-0.000$ & $-0.001$ & $-0.001$ & $-0.002$ & $0.000$ 
		\\
			
			\bottomrule
			\multicolumn{13}{l}{\footnotesize Note: Acc is the proportion of replications with $\hat K=K_0$.} \\
			\multicolumn{13}{l}{\footnotesize MAB is the mean absolute bias of $\hat K$ conditional on incorrect dimension selection.}
		\end{tabular}
	}
\end{table}

\subsection{Comparison with Post-hoc Dimension Selection Criteria}

We further evaluated the dimension selection performance of the proposed method under the true latent dimension \(K_0=3\) by comparing it with several standard post-hoc model selection criteria. Unlike the proposed approach, which infers the effective latent dimension within a single overfitted model, these competing procedures require fitting a sequence of probit graded response models with different fixed candidate dimensions. Specifically, for each candidate dimension \(H\), we fitted the probit GRM with independent \(\mathcal{N}(0,1)\) priors on the loading parameters \(\beta_{jh}\), and then selected the dimension using AIC, BIC, DIC, WAIC, and \(V\)-fold cross-validation with \(V=5\). We also considered a one-standard-error version of cross-validation, denoted by CV-1SE, which selects the smallest dimension whose cross-validated predictive performance is within one standard error of the best-performing model.

Let \(\{\xi^{(s)}\}_{s=1}^S\) denote posterior samples under a fitted model with a fixed candidate dimension, and let \(\hat{\xi}\) be the posterior mean of the model parameters. The log-likelihood evaluated at \(\xi\) is denoted by
\[
\ell(\xi) = \log \Pr(Y \mid \xi)
= \sum_{i=1}^n \sum_{j=1}^q \log \Pr(Y_{ij}\mid \xi),
\]
where \(Y\) is the \(n \times q\) ordinal response matrix. The AIC is defined as
\[
\mathrm{AIC} = -2\ell(\hat{\xi}) + 2d,
\]
where \(d\) is the number of free parameters. For a probit GRM with \(C\) response categories and candidate dimension \(H\), we take
\[
d = nH + qH + q(C-1),
\]
where \(nH\) corresponds to the subject-specific latent scores, \(qH\) to the item loading parameters, and \(q(C-1)\) to the item threshold parameters.

The BIC is computed as
\[
\mathrm{BIC} = -2\ell(\hat{\xi}) + d\log(N_{\mathrm{obs}}),
\]
where
\[
N_{\mathrm{obs}} = nq
\]
is the total number of observed item responses. The DIC is defined as
\[
\mathrm{DIC} = -2\ell(\hat{\xi}) + 2p_{\mathrm{DIC}},
\]
where
\[
p_{\mathrm{DIC}}
=
2\left\{
\ell(\hat{\xi}) -
\frac{1}{S}\sum_{s=1}^S \ell(\xi^{(s)})
\right\}.
\]

We also computed the WAIC using the pointwise likelihood contributions from the ordinal item responses:
\[
\begin{aligned}
	\mathrm{WAIC}
	=
	-2
	\Bigg[
	&
	\sum_{i=1}^n \sum_{j=1}^q
	\log\left\{
	\frac{1}{S}\sum_{s=1}^S
	\Pr(Y_{ij}\mid \xi^{(s)})
	\right\}
	-
	\sum_{i=1}^n \sum_{j=1}^q
	\mathrm{Var}_{s}
	\left\{
	\log \Pr(Y_{ij}\mid \xi^{(s)})
	\right\}
	\Bigg],
\end{aligned}
\]
where the variance terms are computed across the posterior draws
\(\{\xi^{(s)}\}_{s=1}^S\).

For \(V\)-fold cross-validation, we randomly partitioned the \(n\) subjects into \(V=5\) approximately equal-sized folds. For each fold, the subjects in that fold were treated as the test set, while the remaining subjects formed the training set. For each candidate dimension \(H\), the model was fitted using the training data, and predictive probabilities for the held-out responses were evaluated using the posterior mean of the model parameters. The held-out log-likelihood was then computed as
\[
\ell_{\mathrm{test}}
=
\sum_{i \in \mathcal{I}_{\mathrm{test}}}
\sum_{j=1}^q
\log \Pr(Y_{ij}\mid \hat{\xi}_{\mathrm{train}}),
\]
where \(\mathcal{I}_{\mathrm{test}}\) denotes the set of held-out subjects. The standard CV criterion selects the dimension with the largest average held-out log-likelihood across the \(V\) folds. The CV-1SE criterion selects the smallest dimension whose average held-out log-likelihood is within one standard error of the maximum, thereby encouraging a more parsimonious model.

\begin{figure}[htpb]
	\centering
	\begin{subfigure}{0.46\textwidth}
		\centering
		\includegraphics[width=\linewidth]{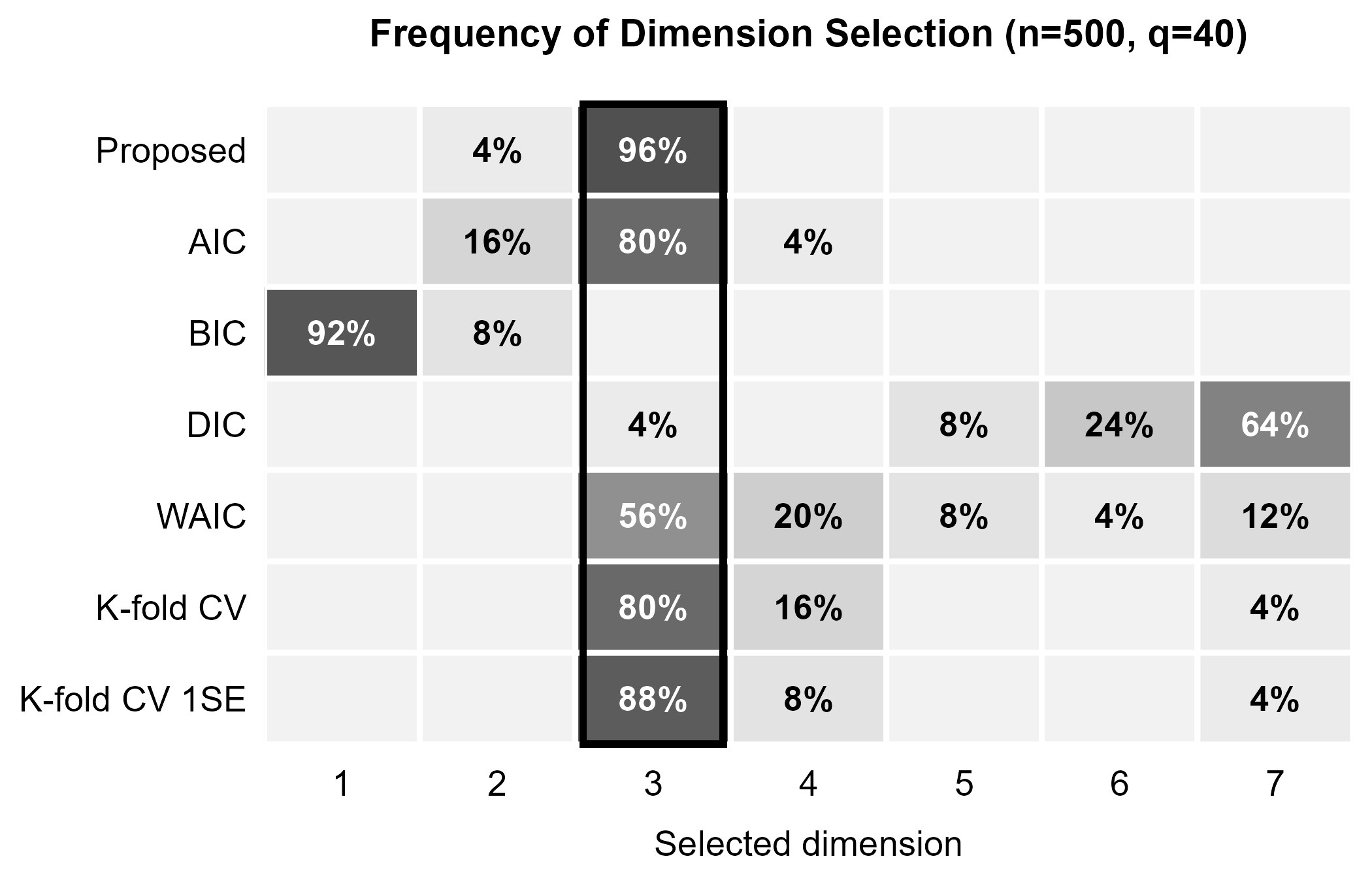}
		\caption*{(a) $n=500, q=40$}
		\label{fig:heat1}
	\end{subfigure}
	\hfill
	\begin{subfigure}{0.46\textwidth}
		\centering
		\includegraphics[width=\linewidth]{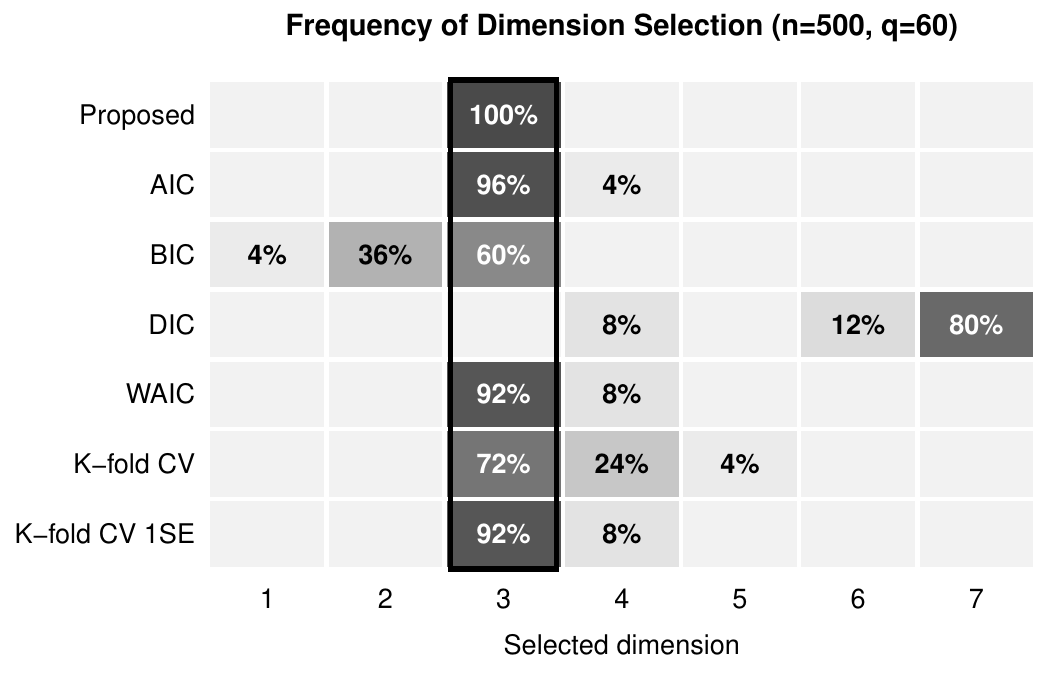}
		\caption*{(b) $n=500, q=60$}
		\label{fig:heat2}
	\end{subfigure}
	
	\vspace{0.5cm}
	
	\begin{subfigure}{0.46\textwidth}
		\centering
		\includegraphics[width=\linewidth]{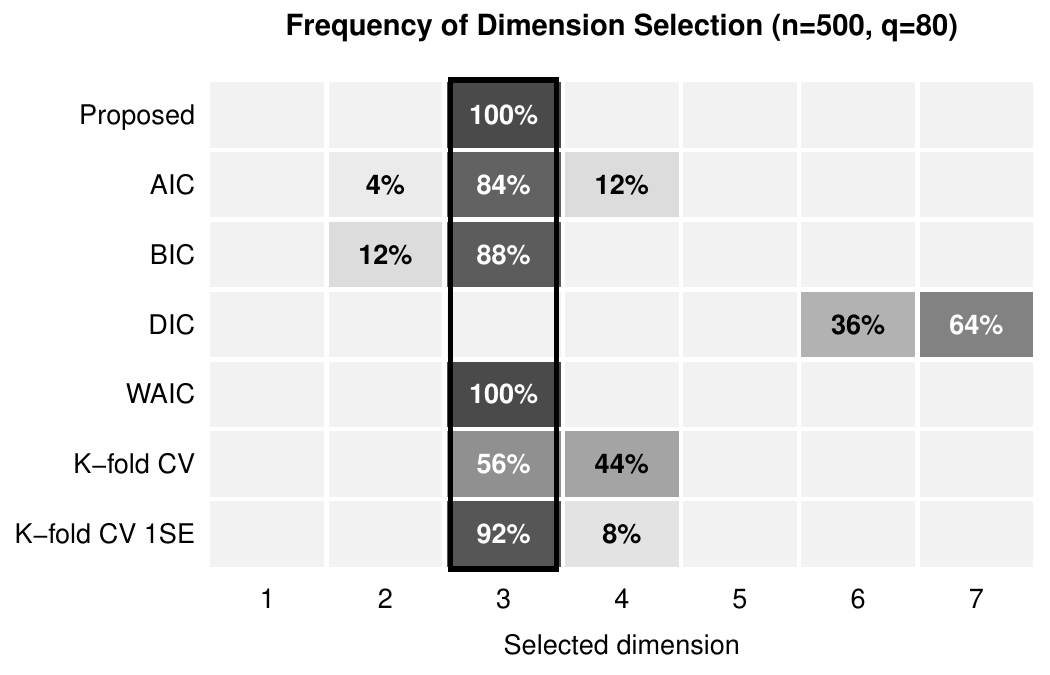}
		\caption*{(c) $n=500, q=80$}
		\label{fig:heat3}
	\end{subfigure}
	\hfill
	\begin{subfigure}{0.46\textwidth}
		\centering
		\includegraphics[width=\linewidth]{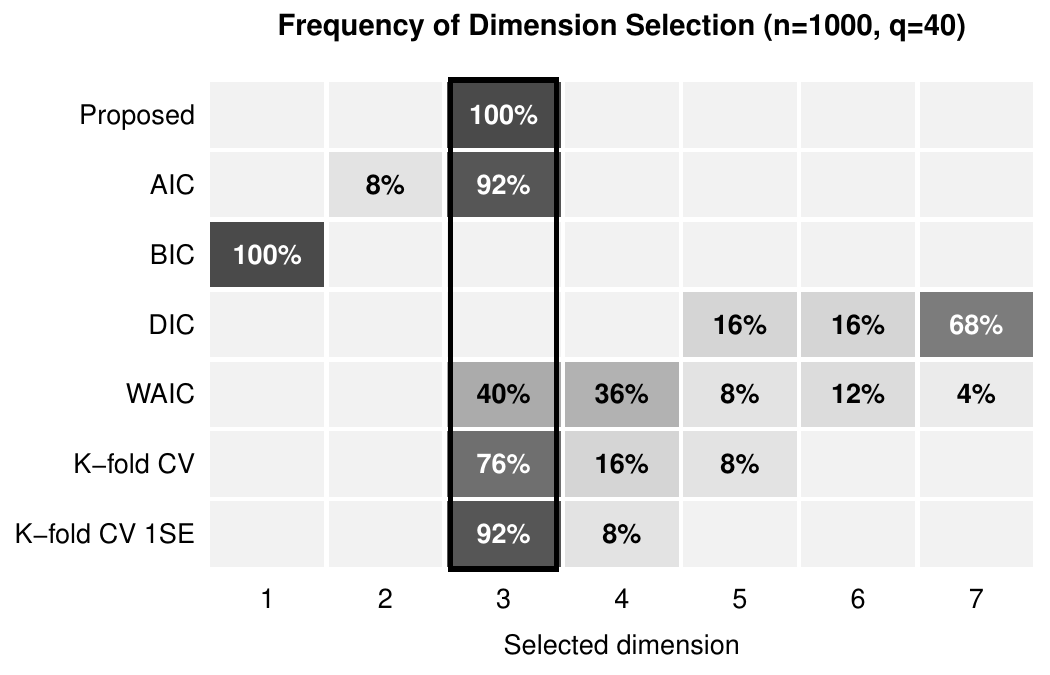}
		\caption*{(d) $n=1000, q=40$}
		\label{fig:heat4}
	\end{subfigure}
	
	\vspace{0.5cm}
	
	\begin{subfigure}{0.46\textwidth}
		\centering
		\includegraphics[width=\linewidth]{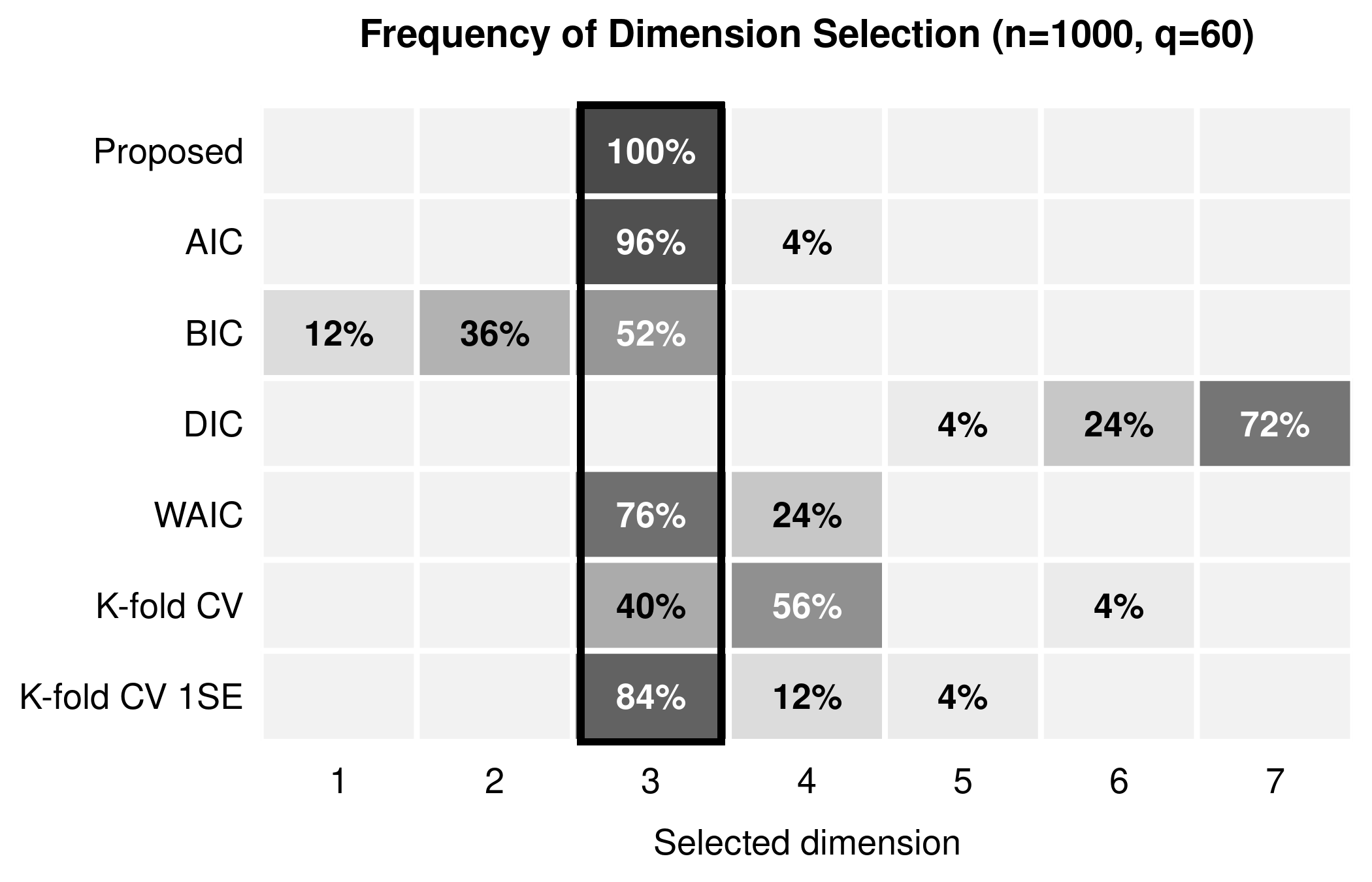}
		\caption*{(e) $n=1000, q=60$}
		\label{fig:heat5}
	\end{subfigure}
	\hfill
	\begin{subfigure}{0.46\textwidth}
		\centering
		\includegraphics[width=\linewidth]{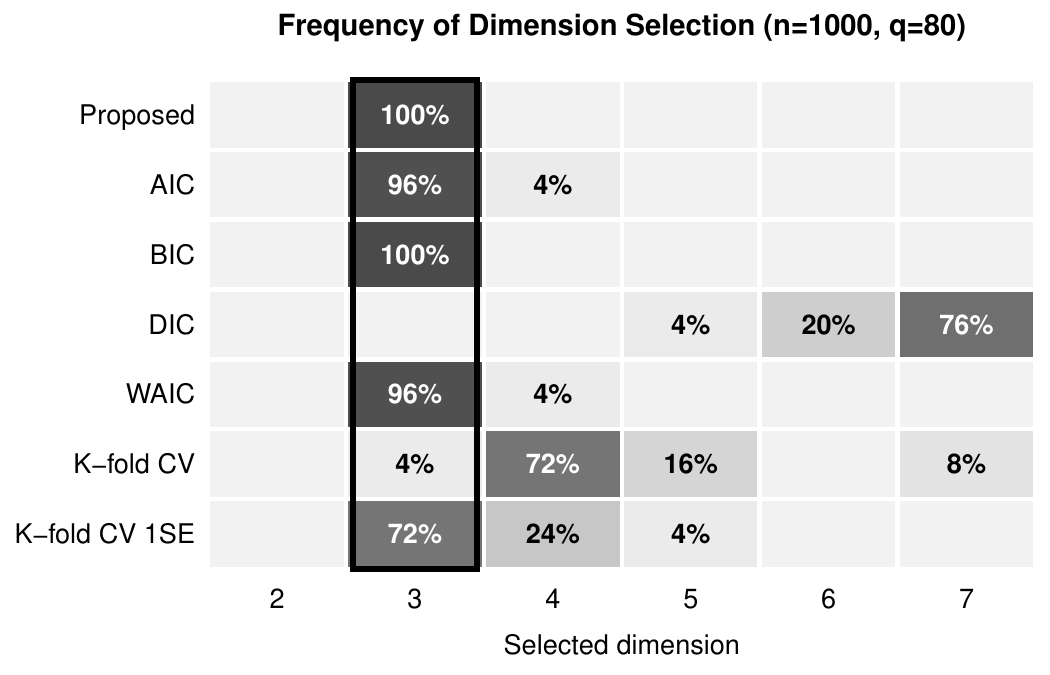}
		\caption*{(f) $n=1000, q=80$}
		\label{fig:heat6}
	\end{subfigure}
	
	\caption{
		Dimension selection performance of the proposed method and competing post-hoc selection criteria for the probit GRM under different configurations of \(n\) and \(q\). Darker colors indicate higher selection frequencies for the corresponding latent dimension.
	}
	\label{fig:heatmaps}
\end{figure}

Figure~\ref{fig:heatmaps} summarizes the dimension selection results across different combinations of \(n\) and \(q\), with the true latent dimension fixed at \(K_0=3\). Correct dimension selection therefore corresponds to selecting \(H=3\). The proposed method achieves high accuracy and remains stable across all considered combinations of sample size and test length. In contrast, the post-hoc criteria exhibit noticeable and systematic biases. DIC and standard cross-validation tend to overestimate the latent dimension, whereas WAIC also shows a tendency to overfit, especially in smaller-sample settings. BIC is more conservative and tends to underestimate the dimension when \(n\) is small, although its performance improves as the sample size increases. AIC performs reasonably well overall, but it is still less stable than the proposed method in smaller samples. The CV-1SE rule mitigates the over-selection problem of standard cross-validation to some extent, but its performance remains sensitive to the sample size and the number of items.

Overall, within the \(K_0=3\) simulation setting considered here, these results demonstrate that the cumulative shrinkage prior provides a robust and accurate mechanism for dimension selection in the probit GRM. Importantly, the proposed approach requires only a single MCMC run under an overfitted model, whereas AIC, BIC, DIC, WAIC, and cross-validation require fitting multiple models over a grid of candidate dimensions. Thus, in addition to improving dimension selection accuracy, the proposed method substantially reduces the computational burden and naturally propagates uncertainty about the latent dimension into posterior inference.

\FloatBarrier
\subsection{Adaptive Estimation versus  Fixed-Dimensional Methods}	
	
Table~\ref{tab:sim_grm_compare} compares the proposed adaptive method with two fixed-dimensional benchmark methods, the Gibbs sampler and the traditional MH-RM algorithm, under the true latent dimension \(K_0=2\). Since both benchmark methods require the latent dimension to be prespecified, they are implemented using the true dimension \(K=K_0\), thereby serving as oracle fixed-dimensional benchmarks. The comparison therefore evaluates whether the proposed method can achieve estimation accuracy comparable to fixed-dimensional oracle methods while retaining the additional ability to recover the latent dimension automatically.

\begin{table}[htbp]
	\centering
	\small
	\setlength{\tabcolsep}{3.2pt}
	\caption{Comparison of the proposed method, the Gibbs sampler, and the MHRM algorithm under different sample sizes and test lengths with true latent dimension \(K_0=2\).}
	\label{tab:sim_grm_compare}
	\resizebox{\textwidth}{!}{
		\begin{tabular}{llcccccccccccc}
			\toprule
			& & \multicolumn{4}{c}{$n=400$} 
			& \multicolumn{4}{c}{$n=800$} 
			& \multicolumn{4}{c}{$n=1200$} \\
			\cmidrule(lr){3-6} \cmidrule(lr){7-10} \cmidrule(lr){11-14}
			Method & Metric 
			& $q=40$ & $q=60$ & $q=80$ & $q=100$
			& $q=40$ & $q=60$ & $q=80$ & $q=100$
			& $q=40$ & $q=60$ & $q=80$ & $q=100$ \\
			\midrule
			
			
			Proposed 
			& $\gamma_{\rm rmse}$ 
			& 0.125 & 0.126 & 0.121 & 0.124 
			& 0.097 & 0.091 & 0.090 & 0.092 
			& 0.075 & 0.076 & 0.074 & 0.075 \\
			
			& $\Delta_B$ 
			& 0.089 & 0.089 & 0.084 & 0.083 
			& 0.062 & 0.061 & 0.057 & 0.059
			& 0.051 & 0.052 & 0.049 & 0.047 \\
			
			& $\Delta_Z$ 
			& 0.604 & 0.520 & 0.458 & 0.410 
			& 0.601 & 0.503 & 0.449 & 0.409 
			& 0.600 & 0.502 & 0.438 & 0.397 \\
			
			& Acc 
			& 100\% & 100\% & 100\% & 100\%
			& 100\% & 100\% & 100\% & 100\%
			& 100\% & 100\% & 100\% & 100\% \\
			
			& MAB 
			& 0.000 & 0.000 & 0.000 & 0.000
			& 0.000 & 0.000 & 0.000 & 0.000
			& 0.000 & 0.000 & 0.000 & 0.000 \\
			
			& $\gamma_{\rm bias}$ 
			& -0.003 & -0.008 & -0.001 & -0.004
			& -0.004 & 0.002 & -0.000 & 0.010
			& -0.002 & 0.008 & -0.004 & 0.000 \\
			
			& $B_{\rm bias}$ 
			& 0.006 & 0.008 & 0.009 & 0.002
			& 0.005 & 0.006 & 0.006 & 0.001
			& 0.005 & 0.004 & 0.008 & 0.003 \\
			
			& $Z_{\rm bias}$ 
			& 0.000 & 0.002 & 0.001 & 0.002
			& -0.001 & -0.001 & 0.001 & 0.000
			& 0.000 & 0.000 & 0.000 & 0.000 \\
			
			\addlinespace[0.3em]
			
			Gibbs 
			& $\gamma_{\rm rmse}$ 
			& 0.124 & 0.122 & 0.121 & 0.122
			& 0.094 & 0.091 & 0.087 & 0.089
			& 0.074 & 0.076 & 0.074 & 0.075 \\
			
			& $\Delta_B$ 
			& 0.097 & 0.106 & 0.117 & 0.121
			& 0.064 & 0.068 & 0.069 & 0.072
			& 0.052 & 0.056 & 0.055 & 0.055 \\
			
			& $\Delta_Z$ 
			& 0.608 & 0.529 & 0.488 & 0.443
			& 0.601 & 0.508 & 0.455 & 0.418
			& 0.601 & 0.506 & 0.442 & 0.403 \\
			
			& Acc 
			& -- & -- & -- & --
			& -- & -- & -- & --
			& -- & -- & -- & -- \\
			
			& MAB 
			& -- & -- & -- & --
			& -- & -- & -- & --
			& -- & -- & -- & -- \\
			
			& $\gamma_{\rm bias}$ 
			& -0.002 & -0.003 & -0.001 & -0.005
			& 0.001 & 0.002 & -0.002 & 0.005
			& -0.009 & 0.003 & 0.001 & -0.005 \\
			
			& $B_{\rm bias}$ 
			& 0.020 & 0.032 & 0.044 & 0.049
			& 0.010 & 0.017 & 0.022 & 0.022
			& 0.008 & 0.010 & 0.018 & 0.017 \\
			
			& $Z_{\rm bias}$ 
			& 0.001 & 0.001 & 0.001 & 0.001
			& -0.001 & 0.000 & -0.000 & -0.001
			& 0.000 & -0.000 & 0.001 & 0.000 \\

			\addlinespace[0.3em]
			
			MHRM 
			& $\gamma_{\rm rmse}$ 
			& 0.291 & 0.297 & 0.308 & 0.327
			& 0.226 & 0.218 & 0.217 & 0.223
			& 0.201 & 0.189 & 0.201 & 0.197 \\
			
			& $\Delta_B$ 
			& 0.124 & 0.130 & 0.105 & 0.102
			& 0.103 & 0.095 & 0.087 & 0.075
			& 0.092 & 0.092 & 0.066 & 0.066 \\
			
			& $\Delta_Z$ 
			& 0.631 & 0.533 & 0.466 & 0.423
			& 0.627 & 0.518 & 0.458 & 0.415
			& 0.625 & 0.513 & 0.447 & 0.407 \\
			
			& Acc 
			& -- & -- & -- & --
			& -- & -- & -- & --
			& -- & -- & -- & -- \\
			
			& MAB 
			& -- & -- & -- & --
			& -- & -- & -- & --
			& -- & -- & -- & -- \\
			
			& $\gamma_{\rm bias}$ 
			& 0.009 & 0.009 & 0.021 &0.023
			& 0.008 & 0.006 & 0.003 & 0.011
			& -0.005 & 0.003 & -0.002 & 0.002 \\
			
			& $B_{\rm bias}$ 
			& 0.025 & 0.019 & 0.015 & -0.000
			& 0.019 & 0.018 & 0.015 & -0.002
			& 0.021 & 0.013 & 0.014 & -0.000 \\
			
			& $Z_{\rm bias}$ 
			& -0.005 & -0.004& -0.004 & -0.004
			& -0.003 & -0.002 & -0.002 & -0.003
			& -0.002 & -0.002 & -0.001 & -0.002 \\

			\bottomrule
			\multicolumn{14}{l}{\footnotesize Note: Gibbs and MHRM are implemented with the true latent dimension $K=K_0$ and therefore serve as an oracle benchmark.} \\
			\multicolumn{14}{l}{\footnotesize Acc is the proportion of replications with $\hat K=K_0$. MAB is the mean absolute bias of $\hat K$ conditional on incorrect dimension selection.} \\
			\multicolumn{14}{l}{\footnotesize Acc and MAB are not applicable to Gibbs and MHRM because the latent dimension is fixed in advance.}
		\end{tabular}
	}
\end{table}

The results show that, when \(K_0=2\), the proposed adaptive method achieves estimation accuracy comparable to, and in several cases better than, the oracle fixed-dimensional benchmarks. In particular, the proposed method yields threshold RMSEs similar to those of the Gibbs sampler and substantially smaller than those of MHRM, while maintaining comparable errors for the loading and latent score structures. Moreover, unlike the fixed-dimensional benchmark methods, the proposed method accurately recovers the latent dimension in all reported settings, as indicated by Acc \(=100\%\) and MAB \(=0\).
	
	\section{Real Data Analysis}

As an illustration, we apply the proposed probit graded response model with adaptive dimension selection to the publicly available IPIP-NEO personality assessment data obtained from \url{https://osf.io/tbmh5/}.
The IPIP-NEO inventory is a public-domain representation of the widely used NEO personality inventory framework \citep{costa1985neo}, and is designed to measure the Big Five personality factors: Neuroticism (N), Extraversion (E), Openness to experience (O), Agreeableness (A), and Conscientiousness (C). The original dataset contains 20,993 respondents and 300 items. In this analysis, we use a complete-case subset of 3,000 respondents who provided valid responses to all 300 items.

The goal of this real-data analysis is twofold. First, we examine whether the proposed algorithm can recover the theoretically expected Big Five structure without fixing the number of latent factors in advance. Second, we assess whether the proposed model can produce interpretable parameter estimates, including the factor loading structure and item-category threshold parameters, 
under a high-dimensional ordinal response setting. For model identifiability, we impose a lower-triangular constraint on the loading matrix, which is sufficient under our theoretical framework while still allowing the effective latent dimension to be learned from the data.

Rather than pre-specifying a five-dimensional structure as is commonly done in confirmatory factor analysis, we fit the model with a conservatively large upper bound $K=10$ on the number of latent dimensions. This overfitted specification allows the cumulative shrinkage prior to shrink redundant dimensions toward 
inactivity and to determine the effective number of factors in a data-adaptive manner. Therefore, the analysis provides a direct empirical test of whether the proposed method can simultaneously select the latent dimensionality and estimate the parameters of a graded response model for large-scale personality assessment 
data.

\begin{figure}[htpb]
	\centering
	\begin{subfigure}{0.46\textwidth}
		\centering
		\includegraphics[width=\linewidth]{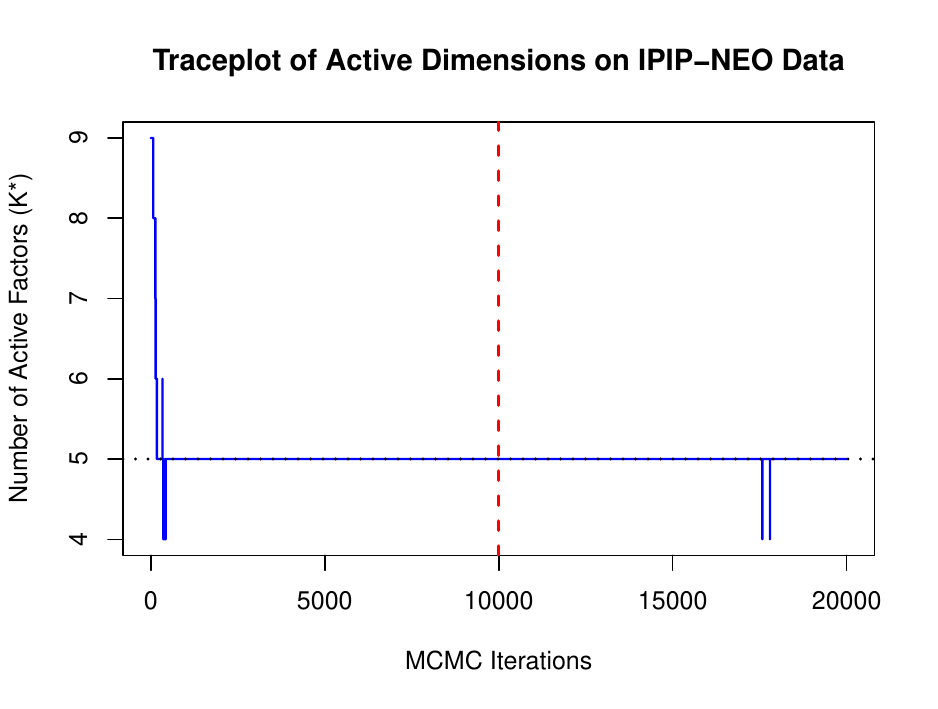}
		\caption{Run 1}
		\label{fig:trace_run1}
	\end{subfigure}
	\hfill
	\begin{subfigure}{0.46\textwidth}
		\centering
		\includegraphics[width=\linewidth]{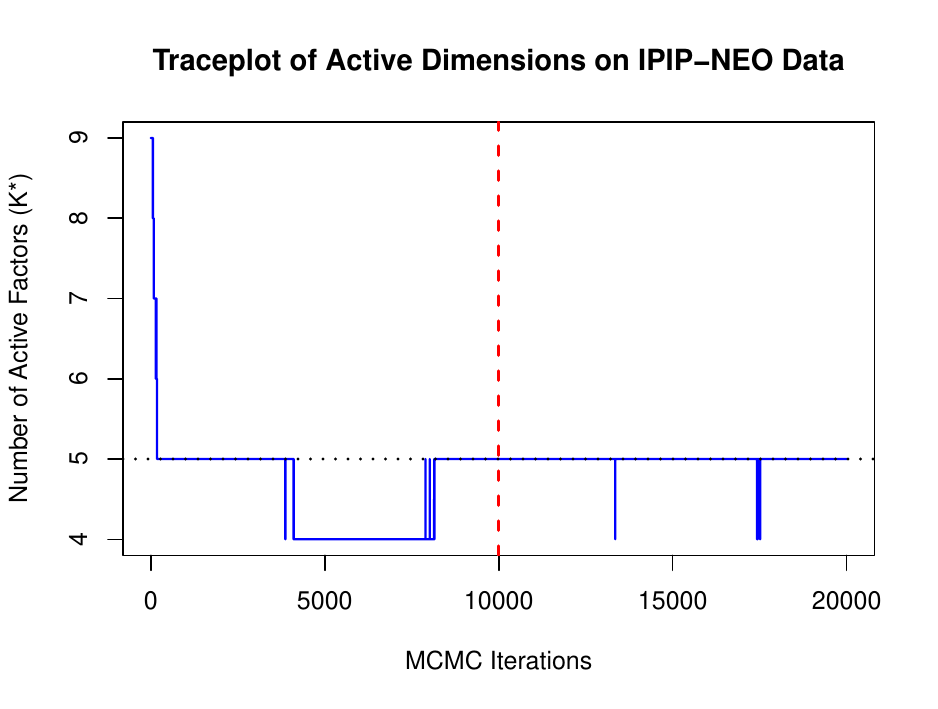}
		\caption{Run 5}
		\label{fig:trace_run5}
	\end{subfigure}
	
	\vspace{0.5cm}
	
	\begin{subfigure}{0.46\textwidth}
		\centering
		\includegraphics[width=\linewidth]{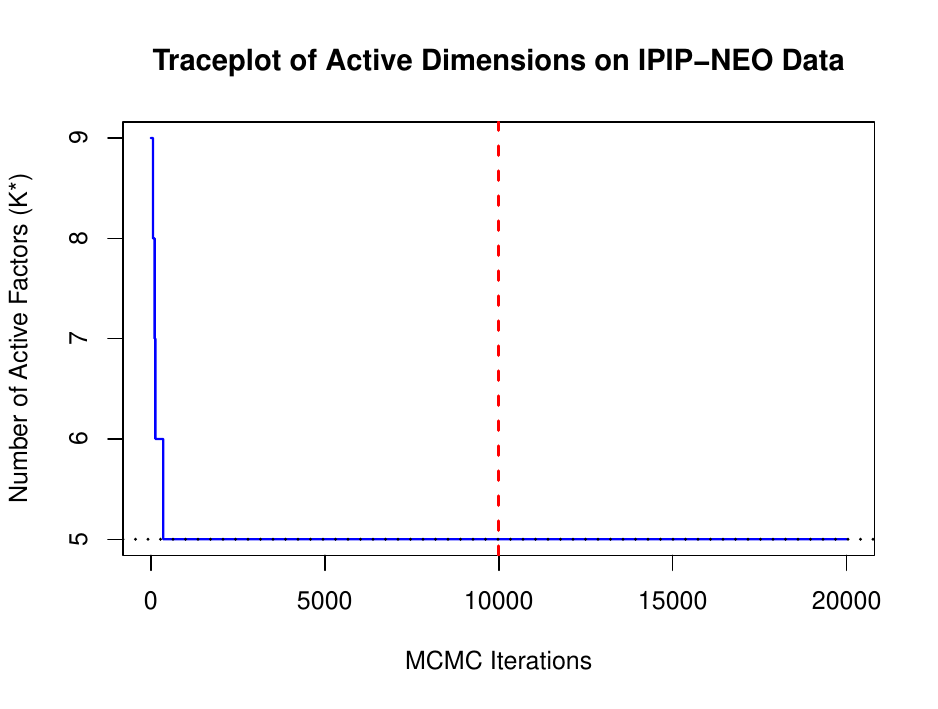}
		\caption{Run 10}
		\label{fig:trace_run10}
	\end{subfigure}
	\hfill
	\begin{subfigure}{0.46\textwidth}
		\centering
		\includegraphics[width=\linewidth]{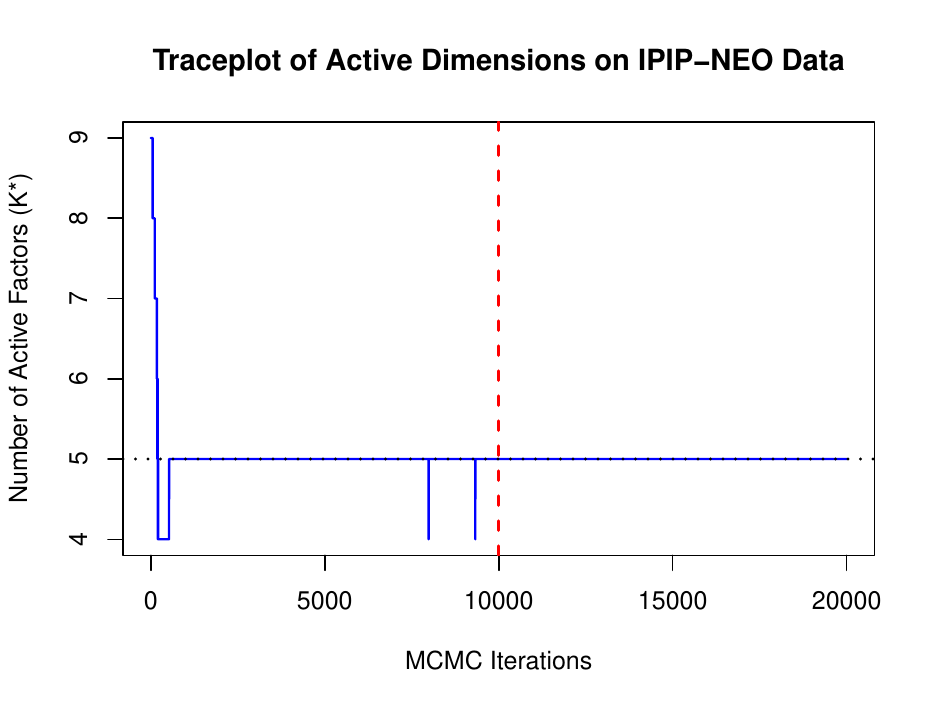}
		\caption{Run 15}
		\label{fig:trace_run15}
	\end{subfigure}
	
	\vspace{0.5cm}
	
	\begin{subfigure}{0.46\textwidth}
		\centering
		\includegraphics[width=\linewidth]{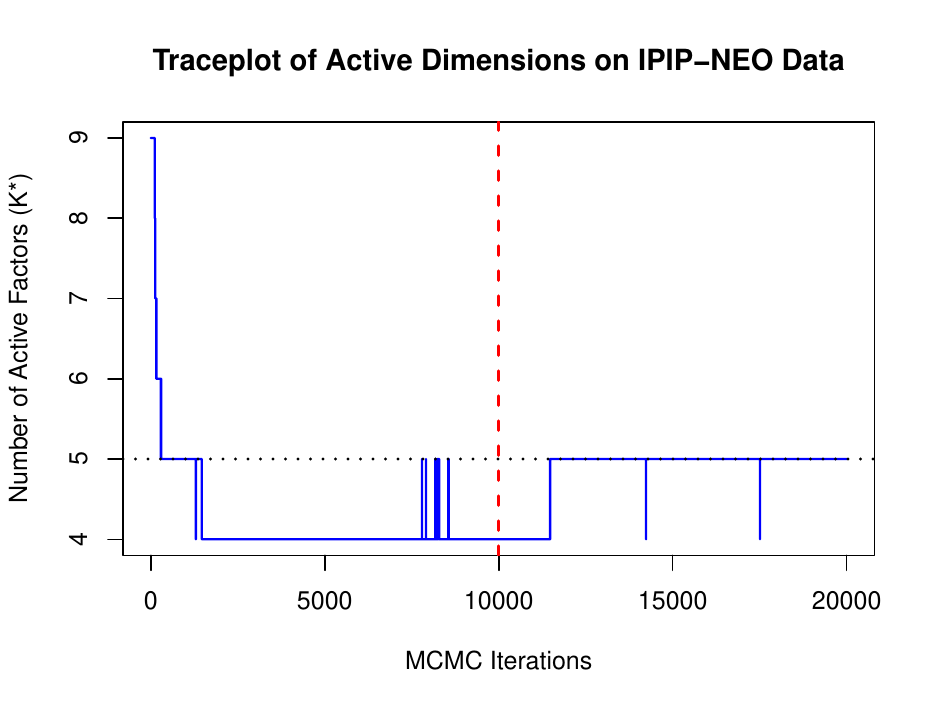}
		\caption{Run 20}
		\label{fig:trace_run20}
	\end{subfigure}
	\hfill
	\begin{subfigure}{0.46\textwidth}
		\centering
		\includegraphics[width=\linewidth]{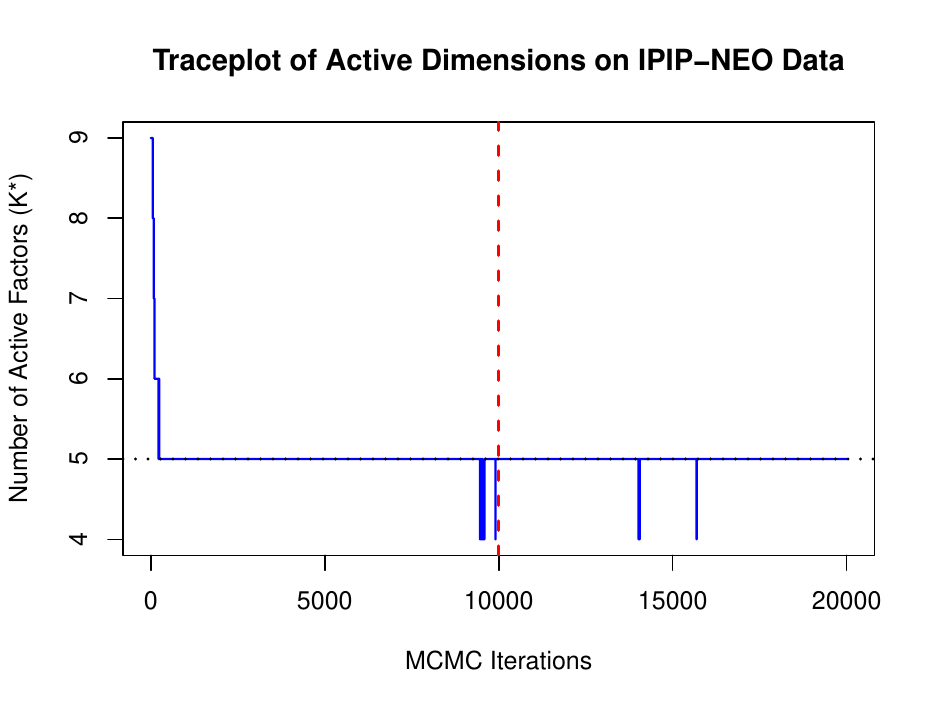}
		\caption{Run 25}
		\label{fig:trace_run25}
	\end{subfigure}
	
	\caption{Traceplots of the number of active dimensions for six representative independent runs on the IPIP-NEO dataset. The selected runs correspond to runs 1, 5, 10, 15, 20, and 25.}
	\label{fig:active_traceplots}
\end{figure}

To assess the stability of dimension selection, we repeated the analysis across 25 independent runs with different random initializations. Across 25 independent runs on the real IPIP-NEO dataset, the proposed algorithm selected the theoretically anticipated Big Five structure, $\hat{K} = 5$, in 24 runs 
(96\%), as summarized in Table~\ref{tab:real_dimension_selection}. Representative traceplots of the number of active dimensions from six independent runs are shown in
Figure~\ref{fig:active_traceplots}. These traceplots indicate that, after an initial transient phase, the sampler typically stabilizes around five active dimensions, providing a visual diagnostic consistent with the numerical selection 
results.

The only deviation was a single run in which the sampler selected $\hat{K}  = 3$, suggesting a rare case of under-extraction, possibly due to convergence to a local posterior mode or residual Monte Carlo variability. Overall, the strong concentration of the selected dimensionality at $\hat{K}  = 5$ provides empirical 
evidence for the robustness of the algorithm in recovering the dominant latent personality structure without requiring the number of factors to be specified in advance.

\begin{table}[ht]
	\centering
	\caption{Dimension selection results across 25 independent runs on the IPIP-NEO dataset.}
	\label{tab:real_dimension_selection}
	\begin{tabular}{ccc}
		\hline
		Selected dimension  & Number of runs & Proportion \\
		\hline
		3 & 1 & 4\% \\
		5 & 24 & 96\% \\
		\hline
	\end{tabular}
\end{table}

After determining the effective latent dimensionality, we further examined the estimated slope parameters to assess the interpretability of the recovered factors. The complete set of estimated primary loadings is reported in Table~\ref{tab:neo_full_loadings}. Each entry gives the posterior mean of the estimated slope parameter, with the corresponding item number shown in parentheses. The loading pattern is broadly consistent with the intended IPIP-NEO item organization: items assigned to the same theoretical domain tend to load most strongly on the corresponding recovered latent trait. To facilitate 
interpretation, Table~\ref{tab:real_top_5} further lists the five items with the largest estimated loadings for each factor.  Overall, the recovered factors show a clear and interpretable correspondence with the Big Five domains. Items with 
the largest loadings on the first factor are primarily related to emotional instability and stress reactivity, corresponding to Neuroticism. The second factor is mainly associated with interpersonal engagement and social withdrawal, 
which is consistent with Extraversion. The third factor is dominated by items reflecting aesthetic sensitivity, reflection, and intellectual curiosity, corresponding to Openness to experience. The fourth factor is characterized by items related to distrust, antagonism, and interpersonal conflict, matching 
Agreeableness after accounting for reverse-keyed item coding. The fifth factor is associated with orderliness, rule-following, effort, and self-discipline, which is consistent with Conscientiousness.

These results indicate that the proposed probit graded response model not only selects the theoretically expected number of latent dimensions, but also yields substantively meaningful loading estimates. In particular, the items with the largest estimated slopes are consistent with the intended psychological content 
of the corresponding IPIP-NEO domains, providing empirical support for the interpretability of the estimated latent structure.

\begin{table}[htbp]
	\centering
	\small  
	\caption{Estimated primary slope parameters for the IPIP-NEO dataset. Each cell reports the estimated loading, with the corresponding item number shown in parentheses.}
	\label{tab:neo_full_loadings}
	\resizebox{\textwidth}{!}{%
		\begin{tabular}{lcccccccccc}
			\hline
			Latent trait & \multicolumn{10}{c}{Estimated loading parameter with item number} \\
			\hline
			& $0.87(\text{I1})$ & $0.97(\text{I6})$ & $0.95(\text{I11})$ & $0.53(\text{I16})$ & $0.27(\text{I21})$ & $0.94(\text{I26})$ & $0.80(\text{I31})$ & $1.01(\text{I36})$ & $0.80(\text{I41})$ & $0.72(\text{I46})$ \\[3pt]  
			& $0.56(\text{I51})$ & $0.73(\text{I56})$ & $0.77(\text{I61})$ & $1.17(\text{I66})$ & $1.04(\text{I71})$ & $0.37(\text{I76})$ & $0.57(\text{I81})$ & $0.81(\text{I86})$ & $1.30(\text{I91})$ & $0.97(\text{I96})$ \\[3pt]
			$\text{N}$ & $0.85(\text{I101})$ & $0.25(\text{I106})$ & $0.34(\text{I111})$ & $0.49(\text{I116})$ & $1.01(\text{I121})$ & $1.05(\text{I126})$ & $1.10(\text{I131})$ & $0.29(\text{I136})$ & $0.21(\text{I141})$ & $1.13(\text{I146})$ \\[3pt]
			& $0.95(\text{I151})$ & $1.06(\text{I156})$ & $0.79(\text{I161})$ & $0.27(\text{I166})$ & $0.28(\text{I171})$ & $0.69(\text{I176})$ & $0.76(\text{I181})$ & $0.86(\text{I186})$ & $0.44(\text{I191})$ & $0.46(\text{I196})$ \\[3pt]
			& $0.30(\text{I201})$ & $0.39(\text{I206})$ & $0.64(\text{I211})$ & $0.97(\text{I216})$ & $0.78(\text{I221})$ & $0.30(\text{I226})$ & $0.32(\text{I231})$ & $0.63(\text{I236})$ & $0.73(\text{I241})$ & $0.86(\text{I246})$ \\[3pt]
			& $0.72(\text{I251})$ & $0.40(\text{I256})$ & $0.21(\text{I261})$ & $0.41(\text{I266})$ & $0.47(\text{I271})$ & $0.66(\text{I276})$ & $0.64(\text{I281})$ & $0.27(\text{I286})$ & $0.23(\text{I291})$ & $0.75(\text{I296})$ \\[6pt]  
			
			& $0.84(\text{I2})$ & $0.93(\text{I7})$ & $0.28(\text{I12})$ & $0.23(\text{I17})$ & $0.57(\text{I22})$ & $0.77(\text{I27})$ & $0.82(\text{I32})$ & $0.92(\text{I37})$ & $0.25(\text{I42})$ & $0.39(\text{I47})$ \\[3pt]
			& $0.33(\text{I52})$ & $0.88(\text{I57})$ & $1.09(\text{I62})$ & $0.75(\text{I67})$ & $0.16(\text{I72})$ & $0.22(\text{I77})$ & $0.42(\text{I82})$ & $0.39(\text{I87})$ & $0.90(\text{I92})$ & $0.71(\text{I97})$ \\[3pt]
			$\text{E}$ & $0.18(\text{I102})$ & $0.03(\text{I107})$ & $0.88(\text{I112})$ & $0.38(\text{I117})$ & $0.66(\text{I122})$ & $0.68(\text{I127})$ & $0.25(\text{I132})$ & $0.07(\text{I137})$ & $0.24(\text{I142})$ & $0.56(\text{I147})$ \\[3pt]
			& $0.87(\text{I152})$ & $0.88(\text{I157})$ & $0.25(\text{I162})$ & $0.02(\text{I167})$ & $0.72(\text{I172})$ & $0.60(\text{I177})$ & $0.80(\text{I182})$ & $0.87(\text{I187})$ & $0.98(\text{I192})$ & $0.14(\text{I197})$ \\[3pt]
			& $0.20(\text{I202})$ & $0.48(\text{I207})$ & $1.05(\text{I212})$ & $0.81(\text{I217})$ & $0.51(\text{I222})$ & $0.02(\text{I227})$ & $0.11(\text{I232})$ & $0.39(\text{I237})$ & $0.51(\text{I242})$ & $0.93(\text{I247})$ \\[3pt]
			& $0.75(\text{I252})$ & $0.03(\text{I257})$ & $0.10(\text{I262})$ & $0.38(\text{I267})$ & $0.99(\text{I272})$ & $0.86(\text{I277})$ & $0.33(\text{I282})$ & $0.17(\text{I287})$ & $0.30(\text{I292})$ & $0.30(\text{I297})$ \\[6pt]
			
			& $1.67(\text{I3})$ & $1.80(\text{I8})$ & $1.71(\text{I13})$ & $0.66(\text{I18})$ & $0.93(\text{I23})$ & $0.44(\text{I28})$ & $1.45(\text{I33})$ & $1.41(\text{I38})$ & $1.98(\text{I43})$ & $1.39(\text{I48})$ \\[3pt]
			& $1.25(\text{I53})$ & $0.41(\text{I58})$ & $1.75(\text{I63})$ & $2.15(\text{I68})$ & $1.19(\text{I73})$ & $1.92(\text{I78})$ & $1.28(\text{I83})$ & $0.21(\text{I88})$ & $2.00(\text{I93})$ & $1.34(\text{I98})$ \\[3pt]
			$\text{O}$ & $1.33(\text{I103})$ & $1.36(\text{I108})$ & $1.35(\text{I113})$ & $0.43(\text{I118})$ & $0.96(\text{I123})$ & $2.40(\text{I128})$ & $2.04(\text{I133})$ & $0.06(\text{I138})$ & $2.43(\text{I143})$ & $0.60(\text{I148})$ \\[3pt]
			& $2.09(\text{I153})$ & $2.11(\text{I158})$ & $1.16(\text{I163})$ & $0.39(\text{I168})$ & $1.64(\text{I173})$ & $1.06(\text{I178})$ & $1.62(\text{I183})$ & $1.68(\text{I188})$ & $1.33(\text{I193})$ & $0.54(\text{I198})$ \\[3pt]
			& $1.67(\text{I203})$ & $0.25(\text{I208})$ & $1.94(\text{I213})$ & $1.58(\text{I218})$ & $1.39(\text{I223})$ & $0.23(\text{I228})$ & $1.74(\text{I233})$ & $0.02(\text{I238})$ & $1.50(\text{I243})$ & $1.00(\text{I248})$ \\[3pt]
			& $1.09(\text{I253})$ & $1.01(\text{I258})$ & $1.84(\text{I263})$ & $0.41(\text{I268})$ & $1.97(\text{I273})$ & $1.12(\text{I278})$ & $1.67(\text{I283})$ & $0.47(\text{I288})$ & $1.58(\text{I293})$ & $0.46(\text{I298})$ \\[6pt]
			
			& $0.36(\text{I4})$ & $0.08(\text{I9})$ & $0.04(\text{I14})$ & $0.34(\text{I19})$ & $0.24(\text{I24})$ & $0.29(\text{I29})$ & $0.39(\text{I34})$ & $0.02(\text{I39})$ & $0.02(\text{I44})$ & $0.48(\text{I49})$ \\[3pt]
			& $0.14(\text{I54})$ & $0.28(\text{I59})$ & $0.38(\text{I64})$ & $0.19(\text{I69})$ & $0.14(\text{I74})$ & $0.45(\text{I79})$ & $0.30(\text{I84})$ & $0.39(\text{I89})$ & $0.24(\text{I94})$ & $0.31(\text{I99})$ \\[3pt]
			$\text{A}$ & $0.30(\text{I104})$ & $0.64(\text{I109})$ & $0.18(\text{I114})$ & $0.30(\text{I119})$ & $0.30(\text{I124})$ & $0.45(\text{I129})$ & $0.48(\text{I134})$ & $0.33(\text{I139})$ & $0.10(\text{I144})$ & $0.17(\text{I149})$ \\[3pt]
			& $0.40(\text{I154})$ & $0.48(\text{I159})$ & $0.47(\text{I164})$ & $0.40(\text{I169})$ & $0.66(\text{I174})$ & $0.33(\text{I179})$ & $0.51(\text{I184})$ & $0.19(\text{I189})$ & $0.23(\text{I194})$ & $0.32(\text{I199})$ \\[3pt]
			& $0.16(\text{I204})$ & $0.49(\text{I209})$ & $0.26(\text{I214})$ & $0.42(\text{I219})$ & $0.31(\text{I224})$ & $0.21(\text{I229})$ & $0.13(\text{I234})$ & $0.35(\text{I239})$ & $0.38(\text{I244})$ & $0.53(\text{I249})$ \\[3pt]
			& $0.51(\text{I254})$ & $0.38(\text{I259})$ & $0.35(\text{I264})$ & $0.21(\text{I269})$ & $0.26(\text{I274})$ & $0.49(\text{I279})$ & $0.22(\text{I284})$ & $0.13(\text{I289})$ & $0.38(\text{I294})$ & $0.37(\text{I299})$ \\[6pt]
			
			& $0.65(\text{I5})$ & $0.79(\text{I10})$ & $0.89(\text{I15})$ & $0.42(\text{I20})$ & $0.78(\text{I25})$ & $0.45(\text{I30})$ & $0.30(\text{I35})$ & $0.70(\text{I40})$ & $0.45(\text{I45})$ & $0.86(\text{I50})$ \\[3pt]
			& $0.62(\text{I55})$ & $0.25(\text{I60})$ & $0.52(\text{I65})$ & $0.43(\text{I70})$ & $0.56(\text{I75})$ & $0.37(\text{I80})$ & $0.79(\text{I85})$ & $0.49(\text{I90})$ & $0.23(\text{I95})$ & $0.89(\text{I100})$ \\[3pt]
			$\text{C}$ & $0.49(\text{I105})$ & $0.33(\text{I110})$ & $0.81(\text{I115})$ & $0.67(\text{I120})$ & $0.05(\text{I125})$ & $0.79(\text{I130})$ & $0.33(\text{I135})$ & $0.50(\text{I140})$ & $0.71(\text{I145})$ & $0.57(\text{I150})$ \\[3pt]
			& $0.47(\text{I155})$ & $0.61(\text{I160})$ & $0.86(\text{I165})$ & $0.32(\text{I170})$ & $0.93(\text{I175})$ & $0.72(\text{I180})$ & $0.19(\text{I185})$ & $0.66(\text{I190})$ & $0.50(\text{I195})$ & $0.21(\text{I200})$ \\[3pt]
			& $0.80(\text{I205})$ & $0.57(\text{I210})$ & $0.10(\text{I215})$ & $0.60(\text{I220})$ & $0.58(\text{I225})$ & $0.35(\text{I230})$ & $0.69(\text{I235})$ & $0.88(\text{I240})$ & $0.05(\text{I245})$ & $0.53(\text{I250})$ \\[3pt]
			& $0.70(\text{I255})$ & $0.77(\text{I260})$ & $0.73(\text{I265})$ & $0.69(\text{I270})$ & $0.50(\text{I275})$ & $0.69(\text{I280})$ & $0.49(\text{I285})$ & $0.70(\text{I290})$ & $0.50(\text{I295})$ & $0.65(\text{I300})$ \\[3pt]
			\hline
		\end{tabular}%
	}
	\par\vspace{4pt}
	\raggedright{\footnotesize Note: Item numbers are shown in parentheses.}
\end{table}

	\begin{table}[htbp]
		\centering
		\footnotesize
		\caption{Top five items with the largest estimated loadings for each recovered factor.}
		\label{tab:real_top_5}
		\begin{tabular}{llcp{8.5cm}}
			\toprule
			Factor & Item & Loading & Content \\
			\midrule
			\multirow{5}{*}{N}
			& 91 (N$+$)  & 1.30 & Get stressed out easily.\\
			& 66 (N$+$)  & 1.17 & Get upset easily.\\
			& 146 (N$+$) & 1.13 & Get overwhelmed by emotions.\\
			& 131 (N$+$) & 1.10 & Have frequent mood swings.\\
			& 156 (N$-$) & 1.06 & Rarely get irritated.\\
			\midrule
			
			\multirow{5}{*}{E}
			& 62 (E$+$)  & 1.09 & Feel comfortable around people.\\
			& 212 (E$-$) & 1.05 & Avoid contacts with others.\\
			& 272 (E$-$) & 0.99 & Keep others at a distance.\\
			& 192 (E$-$) & 0.98 & Keep in the background.\\
			& 247 (E$-$) & 0.93 & Avoid crowds.\\
			\midrule
			
			\multirow{5}{*}{O}
			& 143 (O$+$) & 2.43 & Enjoy thinking about things.\\
			& 128 (O$+$) & 2.40 & Enjoy the beauty of nature.\\
			& 68 (O$+$)  & 2.15 & See beauty in things that others might not notice.\\
			& 158 (O$-$) & 2.11 & Do not like art.\\
			& 153 (O$+$) & 2.09 & Spend time reflecting on things.\\
			\midrule
			
			\multirow{5}{*}{A}
			& 174 (A$-$) & 0.66 & Think highly of myself.\\
			& 109 (A$-$) & 0.64 & Have a sharp tongue.\\
			& 249 (A$-$) & 0.53 & Take advantage of others.\\
			& 184 (A$-$) & 0.51 & Distrust people.\\
			& 254 (A$-$) & 0.51 & Turn my back on others.\\
			\midrule
			
			\multirow{5}{*}{C}
			& 175 (C$-$) & 0.93 & Find it difficult to get down to work.\\
			& 15 (C$+$)  & 0.89 & Try to follow the rules.\\
			& 100 (C$+$) & 0.89 & Love order and regularity.\\
			& 240 (C$-$) & 0.88 & Do crazy things.\\
			& 50 (C$+$)  & 0.86 & Work hard.\\
			\bottomrule
		\end{tabular}
	\end{table}

\begin{figure}[H]
	\centering
	\includegraphics[width=0.8\linewidth]{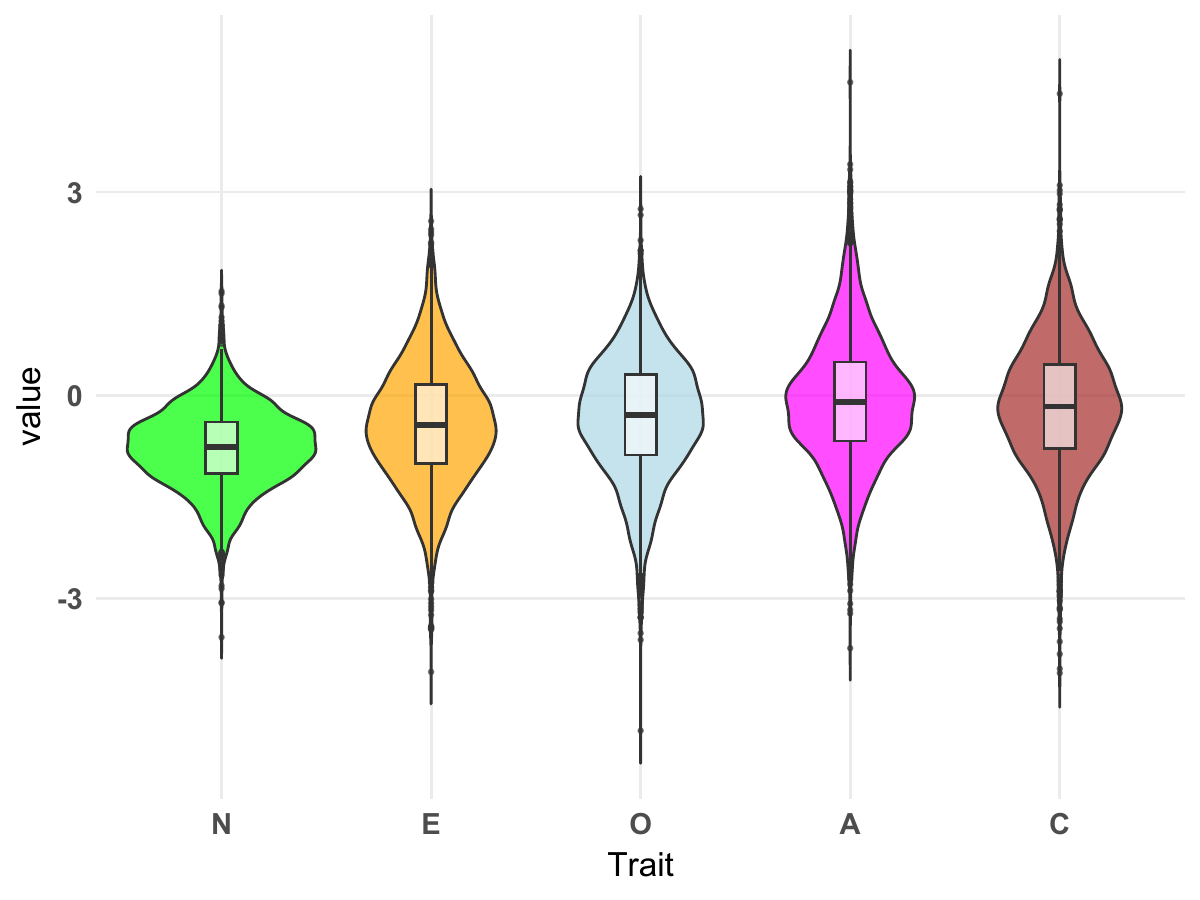}
	\caption{Violin plots of the estimated individual latent factor scores for the five recovered Big Five dimensions in the IPIP-NEO dataset.}
	\label{fig:real_violin_theta}
\end{figure}
In addition to the loading estimates, we also examined the posterior estimates of the individual latent factor scores. Figure~\ref{fig:real_violin_theta} displays violin plots of the estimated latent traits corresponding to the five 
recovered Big Five dimensions. The distributions are centered around zero, which is consistent with the standard location constraint commonly imposed on latent traits, while still showing substantial between-person variability within each 
dimension. This indicates that the proposed model produces non-degenerate and informative estimates of individual-level latent personality traits.

Overall, the estimated factor-score distributions provide an additional diagnostic for the fitted model. Together with the dimension selection results and the interpretable loading patterns, these plots suggest that the proposed COSS-Gibbs sampler can not only recover the effective number of latent dimensions, 
but also estimate meaningful person-level latent traits in the probit graded response model.	
	
\section{Concluding Remarks}
\label{sec:conclusion}

In this paper, we proposed a Bayesian framework for estimating a multidimensional probit graded response model with an unknown latent dimensional structure. The main goal was to provide a flexible yet regularized modeling strategy for ordinal
response data, where the number of latent traits is not fixed in advance but is instead inferred from the data. To achieve this, we introduced a cumulative shrinkage prior on the item loading matrix, which encourages higher-order latent dimensions to be increasingly shrunk toward a spike component. This prior
construction allows the model to retain the most relevant latent dimensions while automatically suppressing redundant ones.

For posterior computation, we developed an adaptive Gibbs sampler based on standard data augmentation for the probit graded response model. The sampler updates the augmented continuous responses, item loading parameters, latent traits, threshold parameters, shrinkage indicators, stick-breaking weights, and
variance parameters in a fully Bayesian manner. To improve computational efficiency, we further incorporated an adaptive truncation strategy. This strategy dynamically removes inactive latent dimensions and adds a terminal spike component when needed, thereby avoiding the need to pre-specify a large fixed truncation level. The adaptation probability decreases over the MCMC iterations, so that adaptation becomes less frequent as the chain evolves.

The proposed method has several practical advantages. First, it avoids treating the latent dimensionality as a fixed tuning parameter and instead estimates the effective number of latent dimensions from the posterior distribution. Second, the cumulative shrinkage structure provides interpretable dimension-specific
regularization, making it possible to distinguish active dimensions from redundant ones. Third, the probit formulation leads to conditionally conjugate updates for many model parameters, which facilitates posterior computation.
Together, these features make the proposed framework useful for exploratory analysis of ordinal response data, especially when the underlying latent structure is complex or uncertain.

There are several directions for future research. First, the current adaptive MGRM framework could be extended to joint latent space models (JLSMs), enabling the simultaneous inference of complex network topologies and ordinal nodal attributes within a shared, self-truncating latent space. Second, as the scale of educational and psychological assessments continues to grow, adapting this exact Gibbs sampler into a "divide-and-conquer" distributed Bayesian framework would significantly enhance its computational scalability for massive datasets. Finally, generalizing the cumulative shrinkage mechanism to accommodate mixed-format assessments (e.g., jointly modeling continuous, nominal, and ordinal responses) or incorporating structured sparsity to account for local item dependence remains an important avenue for future methodological exploration.
	
\section*{Acknowledgements}
	This work is supported by the National Natural Science Foundation of China (Grants Nos. 12271168,12531013, and 12301373) as well as the Shanghai Science and Technology Committee Rising-Star Program (Grant No. 22YF1411100).
	
	\bibliographystyle{apalike}
	\bibliography{refs}

\end{document}